\documentclass[10pt,twocolumn,letterpaper]{article}

\usepackage[pagenumbers]{cvpr} 
\usepackage{graphicx}
\usepackage{capt-of}

\newcommand{\yf}[1]{\textcolor{blue}{#1}}

\newcommand{\hk}[1]{\textcolor{magenta}{#1}}

\usepackage{colortbl}
\usepackage{comment}

\usepackage{multirow}
\usepackage[most, breakable]{tcolorbox}
\usepackage{makecell}

\definecolor{cvprblue}{rgb}{0.21,0.49,0.74}
\usepackage[pagebackref,breaklinks,colorlinks,allcolors=cvprblue]{hyperref}

\def\paperID{43015} 
\def\confName{CVPR}
\def\confYear{2026}

\title{Beyond Single Object: Learning 3D Relations with Large Language Models}

\author{
Kohsuke Ide\textsuperscript{1,2} \quad
Ryousuke Yamada\textsuperscript{1,3} \quad
Yue Qiu\textsuperscript{1} \quad
Xianzheng Ma\textsuperscript{4} \\[1mm]
Yoshihiro Fukuhara\textsuperscript{1} \quad
Hirokatsu Kataoka\textsuperscript{1,4} \quad
Yutaka Satoh\textsuperscript{1,2} \\[2mm]
\textsuperscript{1}AIST \quad
\textsuperscript{2}University of Tsukuba \\
\textsuperscript{3}University of Technology Nuremberg \quad
\textsuperscript{4}University of Oxford
}

\begin{document}

\twocolumn[{
  \renewcommand\twocolumn[1][]{#1}%
  \maketitle
  \begin{center}
    \centering
    \captionsetup{type=figure}
    \includegraphics[width=0.93\textwidth]{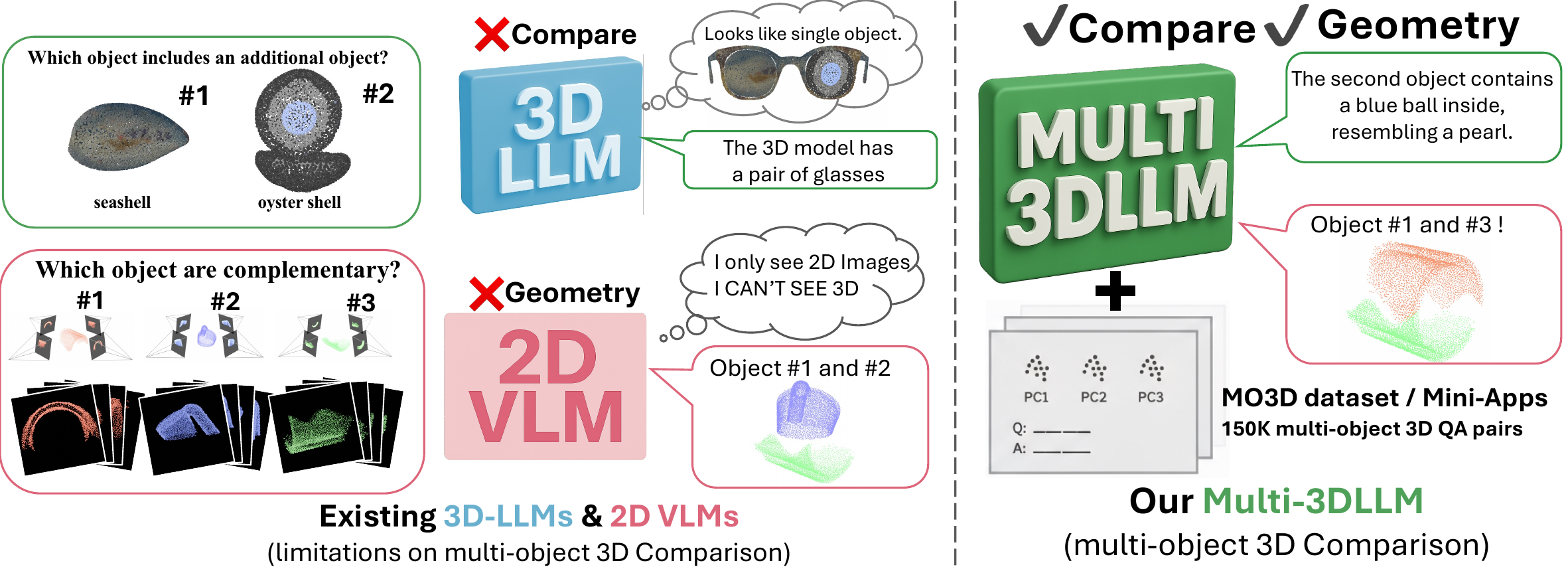}
    \captionof{figure}{
      \textbf{Overview of our proposed multi-object 3D understanding framework.} Left: existing models fail at complex, multi-object 3D tasks. Standard 3D-LLMs cannot perform fine-grained comparisons, and 2D-VLMs are geometrically-unaware, failing tasks like complementarity.
    Right: our Multi-3DLLM, trained on our MO3D and Mini-apps datasets, addresses these challenges, enabling robust comparison (top) and geometric understanding (bottom).
    }
    \label{fig:concept}
  \end{center}
  \vspace{-0pt}
}]


\begin{abstract}
We address a fundamental gap in 3D-LLMs: existing models focus on single-object/scene description, struggling with detailed, inter-object comparison. We propose a framework for detailed object-level reasoning across multiple objects with three components: (1) \textbf{MO3D} (\textbf{M}ulti-\textbf{O}bject in \textbf{3D}), an instruction dataset requiring fine-grained multi-object comparison; (2) Multi-3DLLM, using a minimal Patch-Interaction Transformer (PIT) that models inter-/intra-object relationships while preserving local geometry; (3) Mini-apps, two application-driven benchmarks (Shape Mating, Change Captioning) that probe geometric understanding for practical use. Recent 3D-LLMs and 2D-VLMs perform poorly on these tasks, lacking both comparison-centric design and geometric awareness. In contrast, Multi-3DLLM trained on our mixture data learns geometric reasoning, surpasses all baselines on MO3D, and provides positive transfer to single-object classification.

\end{abstract}


\section{Introduction}
\label{sec:intro}

The emergence of 3D large language models (3D-LLMs) has opened the door to direct interaction with physical space~\cite{xu2024pointllm,qi2024shapellm,qi2024gpt4point}. For the coming era of physical intelligence, however, these models must support not only description but also comparison and relational reasoning over 3D objects~\cite{malisiewicz2009beyond, rosch2024principles}. Existing approaches largely fall into two paradigms: object-centric models that provide high-fidelity descriptions of a single object, and scene-level models~\cite{wang2023chat3d, huang2024chat, huang2023embodied, fu2024scene, chen2024ll3da, abdelreheem2025placeit3d} that process multiple objects for global context. Neither is designed for detailed, object-to-object comparison, which is crucial for applications such as a robot distinguishing similar tools or an augmented reality system comparing furniture. We argue that high-fidelity geometric comparison across distinct point clouds is a critical missing capability for 3D-LLMs.


This capability has been stalled by two key gaps. First, object‑centric 3D‑LLMs such as PointLLM~\cite{xu2024pointllm} and ShapeLLM~\cite{qi2024shapellm} are trained on corpora like Objaverse that consist mainly of isolated objects with no collective or comparative descriptions. Scene‑level models~\cite{wang2023chat3d}, in turn, often rely on object‑level coarse semantics, discarding local geometry needed for precise comparison. As a result, there is no architecture explicitly designed to capture fine‑grained cross‑object relationships, nor a large‑scale instruction dataset that teaches geometric comparison rather than generic scene understanding. Without appropriate data and models, existing methods cannot learn or be evaluated on detailed relationships, differences, and collective properties among multiple 3D objects.

\begin{figure}[t]
    \centering
    \includegraphics[width=0.98\linewidth]{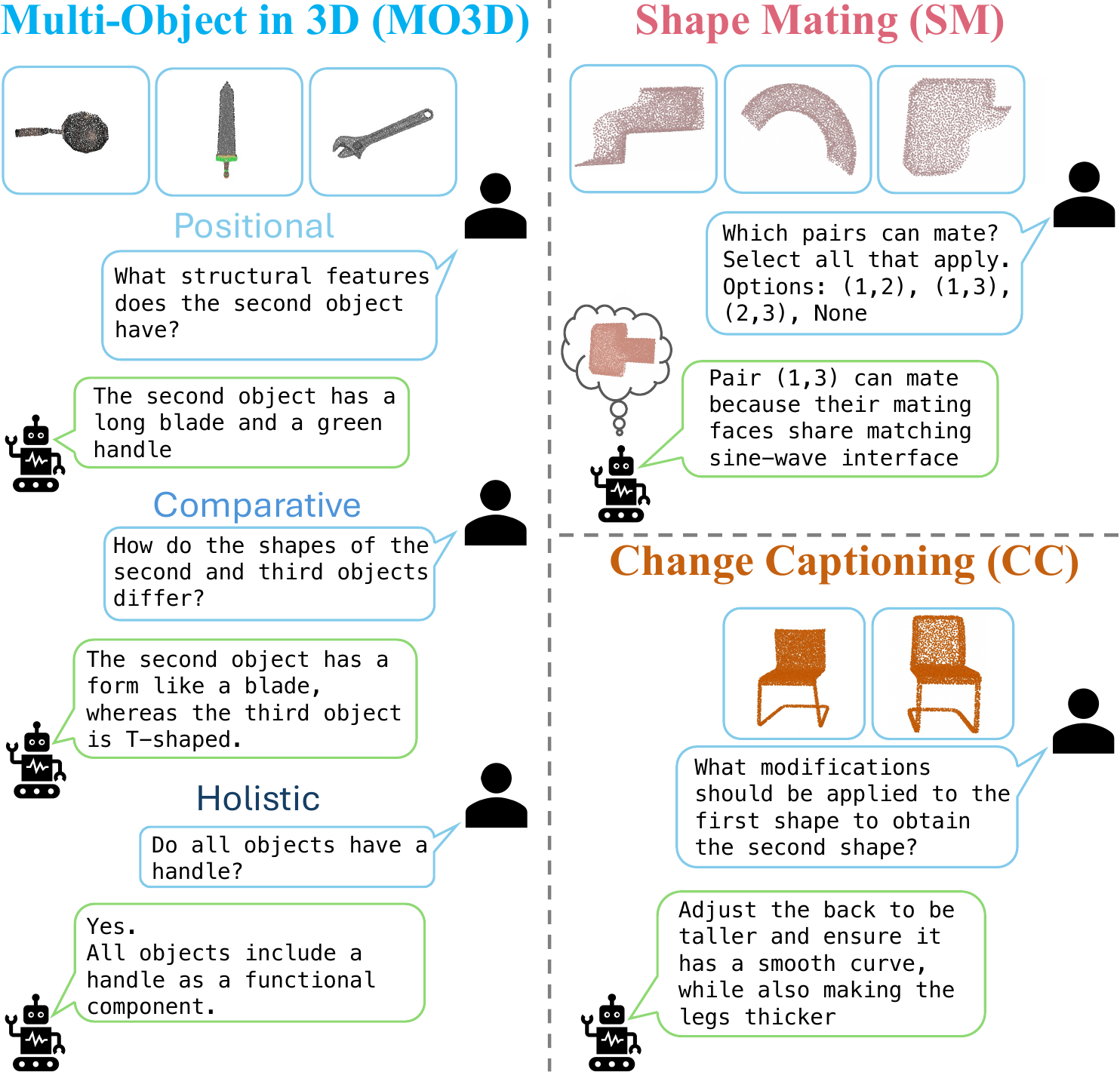}
    \vspace{-10pt}
    \caption{
        \textbf{Overview of our three proposed benchmarks.} 
         MO3D tasks models with multi-object comparison, including \textit{Positional}, \textit{Comparative}, and \textit{Holistic} understanding.
         Shape Mating is a geometrically-intensive task requiring the model to identify the single correct complementary pair from a set of three components.
         Change Captioning tasks the model with generating a natural language description of the geometric edit that transforms an anchor object (left chair) into a target object (right chair).
    }
    \label{fig:mo3d_sm_cc_dataset}
    \vspace{-10pt}
\end{figure}

To address this critical data gap, we introduce MO3D dataset, a large‑scale instruction dataset for multi‑object 3D reasoning, together with two targeted mini‑applications, Shape Mating and Change Captioning (Fig.~\ref{fig:mo3d_sm_cc_dataset}). MO3D dataset forms semantically related sets of point clouds with multiple objects (basically 2 or 3 objects) by grouping Objaverse‑Cap3D~\cite{deitke2023objaverse, luo2023scalable} instances with CLIP~\cite{radford2021learning} embeddings, then prompts an LLM, conditioned on multi‑view renderings and a hierarchy of constraints, to synthesize grounded question–answer pairs. The pipeline explicitly suppresses language priors by balancing question categories, mixing hard positive/negative groupings, and enforcing order invariance, yielding up to 70k positional, comparative, and holistic queries that demand fine‑grained geometric discrimination.

On the modeling side, we propose Multi‑3DLLM, which augments a PointLLM \cite{xu2024pointllm} backbone with a lightweight patch‑interaction transformer. By concatenating patch tokens from all objects and applying a shallow, scalar‑gated self‑attention block, the model captures cross‑object, patch‑level dependencies without destabilizing pretrained alignments or incurring prohibitive cost. This minimal, architecture-agnostic adaptation highlights a key finding: interaction granularity matters. While object-level aggregation smooths details, patch-level processing retains the local geometry required for fine-grained comparative tasks. 

This minimal modification is architecture‑agnostic and illustrates how the broader 3D‑LLM framework can be extended when combined with MO3D dataset.

Extensive experiments reveal that state‑of‑the‑art 3D‑LLMs~\cite{tang2024minigpt3d, qi2024shapellm, tang2025more} and competitive 2D-VLMs~\cite{deitke2025molom, liu2023visual} perform poorly on MO3D’s open‑ended relational tasks, providing empirical evidence showing clear limitations in multi‑object settings. Here, the proposed Multi‑3DLLM trained on our holistic data mixture substantially outperforms the strongest baselines on MO3D dataset, while surpassing models overspecialized to each Mini-App, evidence of generalization rather than memorization. These gains arise from data design: by forcing the model to resolve subtle geometric contrasts instead of relying on textual co‑occurrence, MO3D dataset teaches new relational skills and strengthens foundational ones, notably, zero‑shot single‑object classification improves from 50.7\% to 54.2\%, indicating positive transfer without forgetting. We summarize our contributions as follows.

\noindent\underline{\textbf{Conceptual contribution (Fig.~\ref{fig:concept}).}}

We extend 3D‑LLMs from single‑object or purely scene‑context regimes toward a multi‑object relational paradigm. This enables fine‑grained and multi-object comparison.

\noindent\underline{\textbf{Dataset contribution (Fig.~\ref{fig:mo3d_sm_cc_dataset}).}}
In line with this concept, we construct a suite of benchmarks to greatly expand the capabilities of 3D-LLMs. This includes the MO3D (Multi-Object in 3D) dataset, our instruction dataset for multi-object comparison, along with two critical application-driven benchmarks, Shape Mating and Change Captioning.

\noindent\underline{\textbf{Model contribution (Fig.~\ref{fig:model_architecture}).}} To further unlock the 3D-LLMs potential, we propose Multi-3DLLM. By training with our MO3D dataset, the model can acquire functionalities with minimal modifications to an existing 3D-LLM (\eg, PointLLM). This minimal yet effective contribution demonstrates the extensibility of the 3D-LLM paradigm itself.

\noindent\underline{\textbf{Experimental contribution (Sec.~\ref{sec:experiments}).}}

We provide a comprehensive benchmark showing that existing methods are fundamentally limited on complex, geometry‑aware tasks such as Shape Mating and Change Captioning. Training Multi‑3DLLM on our mixed data yields strong performance on these tasks and produces positive transfer to standard single‑object and point‑wise benchmarks, underscoring the broader applicability of our approach.

\section{Related Work}

\noindent\textbf{Object-Centric 3D-LLMs.}
Research in 3D-LLMs has been largely advanced by an object-centric paradigm~\cite{tang2024minigpt3d, yu2022pointbert, xue2023ulip, xue2024ulip2, qi2024gpt4point, tang2025more, qi2023contrast, pang2023masked, huang2023clip2point}. This line was established by PointLLM~\cite{xu2024pointllm}, which created a direct pipeline from 3D point clouds to LLMs, trained on large-scale caption datasets~\cite{deitke2023objaverse, luo2023scalable}. ShapeLLM~\cite{qi2024shapellm} pursued universal object understanding and embodied interaction by introducing a new 3D encoder, and MiniGPT-3D~\cite{tang2024minigpt3d} demonstrated training efficiency using 2D knowledge. A different paradigm focuses on pre-alignment with unified multi-modal embeddings~\cite{guo2023point, liu2023openshape}. While these models excel at high-fidelity description and recognition~\cite{qi2024gpt4point} of an object, their models and training data confine them to this single-object paradigm. They are not designed to process or compare multiple distinct objects simultaneously.

\noindent\textbf{Scene-Level 3D-LLMs.}
A separate line of work has focused on scene-level understanding, processing multiple objects within a full scene context~\cite{chen2020scanrefer, azuma2022scanqa, hong20233d, li2023m3dbench, yin2023lamm}. Models in this category~\cite{wang2023chat3d, huang2024chat, huang2023embodied, fu2024scene, chen2024ll3da, abdelreheem2025placeit3d} handle multiple objects, typically by aggregating points into coarse object-level tokens to capture global context. While effective for spatial and contextual relationships~\cite{ma2022sqa3d, wald2020learning, zhu20233d, qian2024nuscenes}, this object-level pooling smooths out the local, patch-level details required for intrinsic geometric comparison~\cite{chen2022neural, geng2023gapartnet, zakka2020form2fit}. Consequently, they struggle with the fine-grained geometric analysis our tasks require. Our work addresses the gap between the two paradigms: the patch-level fidelity of object-centric models and the multi-object capability of scene-level models.

\noindent{\textbf{Multi-image and Multi-entity VLMs.}}
The trajectory of 2D-VLMs provides a compelling blueprint for the advancement of 3D understanding. The field experienced a paradigm shift with instruction-tuning, exemplified by InstructBLIP~\cite{dai2023instructblip}, which unlocked a remarkable ability to follow diverse natural language instructions. Flamingo~\cite{alayrac2022flamingo}, Qwen-VL~\cite{bai2023qwen} additionally demonstrated the ability to compare multiple visual inputs. 

The concept of ``comparison'' also has historical precedent in 2D tasks like change captioning~\cite{park2019robust, jhamtani2018learning} and compositional reasoning~\cite{suhr2019nlvr2, suhr2019corpus}. This evolution clearly indicates that enabling multi-entity input is key to unlocking a higher order of understanding. However, this 2D blueprint is often insufficient for tasks demanding strict 3D geometric fidelity. While 2D-VLMs excel at semantic or contextual comparison from projections~\cite{dai2023instructblip, alayrac2022flamingo, bai2023qwen, yang2024qwen2technicalreport, deitke2025molom, liu2023visual}, they lack direct access to the underlying 3D structure~\cite{liu20253daxisprompt, zheng2025learning, huang2025mllms}. A core part of our experimental contribution (Sec.~\ref{sec:experiments}) is to empirically demonstrate that 2D-VLMs, despite their strong language and semantic priors, struggle with the fine-grained, geometry-centric tasks that our 3D-native framework is designed to solve.

\section{MO3D: Multi-Object in 3D Dataset}
\label{sec:dataset}

To extend the capabilities of 3D-LLMs from single-object description to multi-object comparison, we introduce MO3D (Multi-Object in 3D) dataset, an instruction-tuning dataset designed to support both training and evaluation of relational understanding by multi-object comparison, shape mating, and change captioning. The dataset is built on Objaverse-Cap3D~\cite{deitke2023objaverse, luo2023scalable}, which pairs 3D objects from the Objaverse with detailed captions generated by Cap3D. Our pipeline generates approximately 70k high-quality QA pairs, each involving 2-3 point clouds.

MO3D dataset construction follows a two-stage pipeline. First, we use a \textit{Qwen2-72B-Instruct~\cite{yang2024qwen2technicalreport}} to extract naturally occurring attributes from the captions and manually organize them into six core comparison categories. Next, we sample semantically related object groups and generate QA pairs using \textit{GPT-4}. Critically, to ensure questions are grounded in visual facts beyond just the text, we provide \textit{GPT-4} with both the Cap3D captions and multi-view renderings of a point cloud as input. This process targets three types of relational tasks: positional, comparative, and holistic.

\noindent\textbf{Grouping.} We sample groups of 2-3 objects based on the semantic similarity of their captions. We employ a dual sampling inspired by the complementary pair approach~\cite{goyal2017making} to create a diverse mix of scenarios, including both fine-grained comparisons (\eg, three chairs) and broader comparisons involving semantically dissimilar items. To ensure topical diversity, we pre-define six comparison categories (geometry, taxonomy, function, material, style, color) and use weighted sampling to balance their distribution.

\noindent\textbf{QA Generation.} We use an LLM to generate QA pairs for each group. Critically, to ensure questions are grounded in visual facts beyond just the text, we provide the language model with both the Cap3D \cite{luo2023scalable} captions and multi-view renderings of the point clouds as input. This process is guided by a strict prompt hierarchy that enforces visual grounding and ensures clear, unambiguous question formulation. This pipeline targets three core task types. 

\begin{itemize}
    \item \textit{Positional understanding} requires the model to address objects based on their relative order. It requires context-dependent referencing according to each object’s role.
    \item \textit{Comparative understanding} focuses on comparing the properties of multiple objects. It involves analyzing 3D properties like shape, structure, and spatial features.
    \item \textit{Holistic understanding} demands a binary Yes/No judgment that synthesizes information across the entire object set. As the highest level in the hierarchy, it assesses the ability to grasp the whole set and perform reasoning.
\end{itemize}
These three tasks (positional, comparative, and holistic) are not a random assortment. They form a coherent and structured learning curriculum designed to progressively develop sophisticated comparison capabilities.

\noindent{\textbf{Dataset Statistics and Quality Control.}}
\label{ssec:stats}

The final 70k examples are split into training and test sets. Thanks to our balanced sampling, the `holistic' subset features a near 50:50 Yes/No ratio, mitigating linguistic bias for binary questions. To ensure data reliability, we conducted a human audit of 500 randomly sampled QA pairs from the test set. Three independent annotators evaluated each pair, achieving an 81.0\% unanimous agreement rate. Disagreements largely stemmed from intrinsic ambiguities, such as subjective perceptions of abstract shapes or occlusions (detailed in Supp.).


\subsection{Mini-Apps: Application-Driven Benchmarks}
\label{ssec:miniapps}
Beyond MO3D, we curate two application-driven benchmarks, namely shape mating and change captioning.

\noindent{\textbf{Mini-App A: Shape Mating (SM).}}
\label{ssec:shape_mating}
While MO3D focuses on semantic comparison, we introduce the SM benchmark as a downstream task centered on geometric compatibility.

This task is motivated by applications in robotics and industrial design, where determining whether components can physically fit together is crucial. Given three component point clouds, the model must (1) identify which pair, if any, can successfully mate and (2) provide a geometric rationale explaining its decision.
The benchmark is built from meshes sourced from Thingi10K~\cite{zhou2016thingi10k}, using the procedural mesh-cutting process from Neural Shape Mating~\cite{chen2022neural} to split objects into complementary ``part A'' and ``part B'' components. This provides a clear ground truth for mating pairs. We create challenging 4-choice scenarios by sampling a ground-truth and other three incompatible parts.

\noindent{\textbf{Mini-App B: Change Captioning (CC).}
\label{ssec:change_captioning}
Our second mini-application, CC, serves as a benchmark for grounding geometric edits in language. This capability is critical for 3D software applications. The dataset is derived from the ShapeNet~\cite{chang2015shapenet} variation of ShapeTalk dataset~\cite{achlioptas2023shapetalk}, which pairs 3D models with instructions describing the edit.
This benchmark is built upon a contrastive quadruplet, a triplet of point clouds paired with a human-written instruction. The point clouds consist of an \textit{Anchor}, which is the original, pre-edit point cloud, a \textit{Positive}, post-edit point cloud that matches the instruction, and a \textit{Negative}, a distractor that does not follow the Instruction. The \textit{Instruction} itself describes the transformation from the \textit{Anchor} to the \textit{Positive} (e.g., ``thicken the legs''). To create a challenging scenario, the \textit{Negative} model is deliberately sampled to be confusing, originating from the same Anchor but corresponding to a different edit instruction (\eg, ``shorten the legs'').

Based on these components, we generate two distinct task variations of Verification (Binary Classification) and Delta Captioning (Generative). 
For Verification, the model is given the \textit{Anchor}, the \textit{Instruction}, and either the \textit{Positive} or the \textit{Negative} model. It must then answer Yes/No to whether the candidate correctly satisfies the instruction. 
For Delta Captioning, the model is given the \textit{Anchor} and the \textit{Positive} models and is asked to generate a natural language description of the geometric edits that occurred. 

\section{Multi-3DLLM Architecture}
\label{sec:model_architecture}



To validate our data-driven paradigm, we extend PointLLM~\cite{xu2024pointllm}. Originally built for single objects, we modify it to process a set of $N$ objects $O = \{o_1, \ldots, o_N\}$ (Fig.~\ref{fig:model_architecture}). Each object $o_i$ is independently encoded by $g(\cdot)$ and mapped by projector $f(\cdot)$ into a token sequence $Z_i = f\!\big(g(o_i)\big) \in \mathbb{R}^{T \times d}$, where $T$ is the number of patch tokens and $d$ is the LLM dimension. We concatenate these to form a set-wide representation: \begin{equation}
  Z_{\text{cat}} = \text{Concat}(Z_1, Z_2, \ldots, Z_N) \in \mathbb{R}^{(N T) \times d}.
\end{equation}

\paragraph{Patch-Interaction Transformer (PIT) block.}


We insert a minimal Transformer encoder $F_\theta$ performing self-attention jointly over all patch tokens. We use a scalar-gated residual: 
\begin{equation}
    \hat{Z}_{\text{cat}} \;=\; Z_{\text{cat}} + \gamma \,\Delta
    \;=\; (1-\gamma)\,Z_{\text{cat}} + \gamma\,F_\theta(Z_{\text{cat}}).
\end{equation}
This lets $\gamma$ control the PIT contribution. Crucially, initializing $\gamma \approx 0$ preserves pre-trained alignments, allowing the model to safely learn cross-object relations without catastrophic forgetting of foundational single-object knowledge.

\begin{figure*}[t]
    \centering
    \captionsetup{font=small,skip=3pt}
    \includegraphics[width=0.95\textwidth]{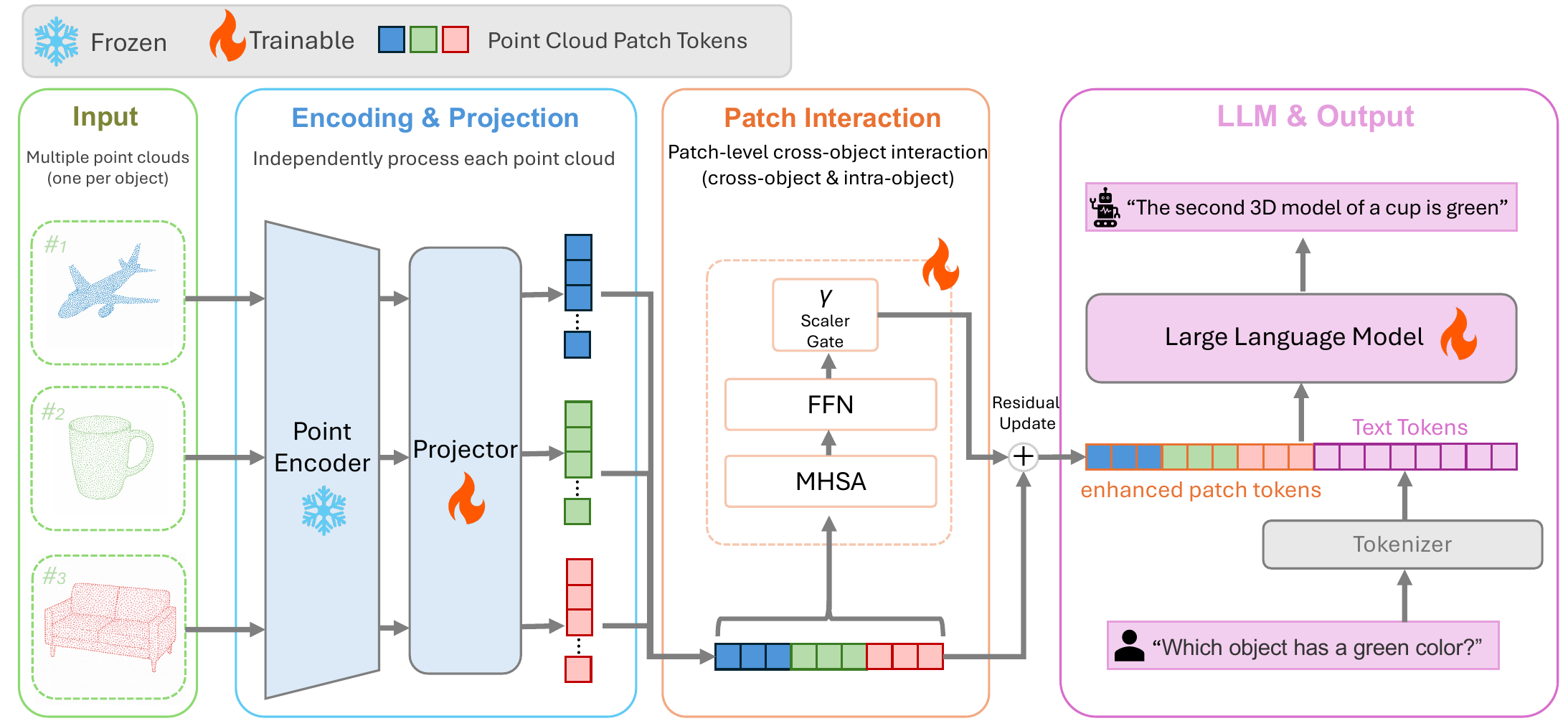}
    \caption{\textbf{Multi-3DLLM architecture.}
$N$ point clouds are encoded independently by the Point Encoder and Projector.
A PIT Block then performs self-attention over the concatenated point tokens, injecting cross-object relations before the tokens   are passed to the LLM.}
    \label{fig:model_architecture}
    \vspace{-8pt}
\end{figure*}


This representation $\hat{Z}_{\text{cat}}$ reflects our core design principle: \textit{interaction granularity matters}. Unlike prior scene-level work~\cite{wang2023chat3d} that compresses each object into a single vector, our patch-level interaction preserves the fine-grained local geometry required for complex geometric tasks. This formulation endows the model with multi-object reasoning capabilities, motivating our instruction-tuning (Sec.~\ref{ssec:training_strategy}).

\subsection{Training Strategy}
\label{ssec:training_strategy}
Training procedure for our Multi-3DLLM is structured into two distinct phases, building upon the two-stage strategy of PointLLM~\cite{xu2024pointllm}. The objective is to first align the core feature representations, then adapt the model to a multi-object context, and finally, teach relational reasoning skills.

\noindent{\textbf{Phase 1: Feature Alignment.}} We begin by inheriting the methodology from PointLLM's first training stage. The parameters of the point cloud encoder and the LLM are frozen, and only the MLP projector is trained. This phase uses the large-scale, brief-description instruction dataset to effectively align the features of point clouds with the LLM's text embedding space.

\noindent{\textbf{Phase 2: Holistic Task-Mixture Fine-tuning.}}
We unfreeze the MLP projector, PIT block, and LLM, and instruction-tune on a unified mixture dataset \textbf{of $\approx$150K pairs}, sampled from \textbf{the MO3D train split} ($\approx$63K), SM ($\approx$44K), and CC ($\approx$40K). \textbf{Note that MO3D contains $\approx$70K pairs in total across all splits.} This end-to-end mixed-task training lets the projection layer and PIT block learn representations directly optimized for our target tasks, in contrast to methods that pre-train separate components~\cite{wang2023chat3d} on potentially out-of-domain data, which require more data and time and can trap the projector in a sub-optimal, poorly aligned solution. Across both phases, we minimize the negative log-likelihood of the response tokens, enabling the model to effectively integrate geometric, linguistic, and relational information.

\begin{table*}[t]
\centering
\caption{
     Comprehensive comparison of Multi-3DLLM against SOTA 3D-LLM and 2D-VLM baselines across all proposed tasks: MO3D, Shape Mating, and Change Captioning. 
    Metrics are reported as: M = Semantic Accuracy, B = Binary Accuracy, R = Reasoning Accuracy, and S = Selection Accuracy (all in \%).
    For 3D-LLM baselines, ``Concat.'' denotes the point cloud concatenation strategy. 
    For 2D-VLM baselines, ``(n images)'' and ``(2*n images)'' refer to providing 1 view and 2 opposing views per object, respectively.
}
\vspace{-0pt}
\label{tab:pilot-results}
\resizebox{\textwidth}{!}{%
\begin{tabular}{l | c | c | cc | cc | cc | c}
\toprule
\multirow{4}{*}{\textbf{Model}} 
& \multicolumn{4}{c|}{\textbf{MO3D}} 
& \multicolumn{2}{c|}{\textbf{Shape Mating}} 
& \multicolumn{3}{c}{\textbf{Change Captioning}} \\ 
\cmidrule(lr){2-5} \cmidrule(lr){6-7} \cmidrule(lr){8-10} 
& \multicolumn{1}{c|}{\textbf{Positional (\%)}} 
& \multicolumn{1}{c|}{\textbf{Comparative (\%)}} 
& \multicolumn{2}{c|}{\textbf{Holistic (\%)}} 
& \multicolumn{2}{c|}{\textbf{Selection (\%)}}
& \multicolumn{2}{c|}{\textbf{Verify (\%)}}   
& \multicolumn{1}{c}{\textbf{Delta Caption (\%)}} \\

& M & M & B & R & S & R & B & R & M \\
\midrule

LLaVA-7B (n images) 
& 25.8 & 11.7 & 61.1 & 41.9 & 20.4 & 2.0 & 48.2 & 29.6 & 7.6 \\

LLaVA-7B (2*n images) 
& 21.3 & 4.8 & 59.9 & 31.0 & 21.9 & 2.5 & 49.1 & 34.6 & 9.3 \\

Molmo-7B (n images) 
& 16.5 & 10.5 & 61.8 & 41.6 & 20.7 & 0.2 & 27.8 & 29.6 & 0.0 \\

Molmo-7B (2*n images) 
& 16.8 & 6.9 & 63.3 & 36.5 & 22.7 & 0.3 & 34.3 & 28.1 & 0.0 \\
\midrule

PointLLM-7B (Concat.) 
& 15.6 & 4.2 & 66.0 & 33.4 & 21.4 & 1.7 & 42.7 & 33.8 & 0.4 \\
 
ShapeLLM-7B (Concat.) 
& 22.6 & 11.2 & 49.8 & 34.6 & 17.5 & 0.1 & 47.2 & 43.2 & 1.0 \\
 
MiniGPT-3D (Concat.) 
& 20.1 & 7.8 & 62.4 & 35.1 & 22.2 & 0.1 & 4.8 & 1.2 & 0.4 \\
 
\midrule
Multi-3DLLM (No-Interaction) 
& 45.5 & 21.8 & 81.3 & 53.0 & 34.4 & 36.7 & 49.1 & 37.1 & 48.0 \\

Multi-3DLLM (Object-Level)
& 52.9 & 32.3 & 81.0 & 49.7 & 25.0 & 23.7 & \textbf{51.7} & 34.7 & 49.6 \\
 
\rowcolor[gray]{0.7}\textbf{Multi-3DLLM (Ours)} 
& \textbf{56.3} & \textbf{33.8} & \textbf{81.7} & \textbf{57.2} & \textbf{37.1} & \textbf{36.8} & 51.2 & \textbf{37.3} & \textbf{51.0} \\
\bottomrule
\end{tabular}
}
\vspace{-10pt}
\end{table*}

\begin{table}[t]
\centering
\small
\setlength{\tabcolsep}{3pt}
\caption{
Zero-shot classification (ModelNet40) results for single-object (1-Obj) versus multi-object (Multi-Obj) inputs. `M' denotes the ordinal position of the queried object in 2-input and 3-input scenes, `MI' indicates multiple 3D point cloud inputs, and `Ours' employs Multi-3DLLM architecture.
}
\vspace{-0pt}
\label{tab:mo3d_adapt_results}
\begin{tabular}{l|c|cc|ccc}
\toprule
& \multicolumn{1}{c|}{\textbf{1-Obj}} & \multicolumn{5}{c}{\textbf{Multi-Obj Task (\%)}} \\
\cmidrule(lr){3-7}
 & \textbf{Task} & \multicolumn{2}{c|}{\textbf{2 Inputs}} & \multicolumn{3}{c}{\textbf{3 Inputs}} \\
\textbf{Model} & \textbf{(\%)} & \textbf{M=1} & \textbf{M=2} & \textbf{M=1} & \textbf{M=2} & \textbf{M=3} \\
\midrule
PointLLM~\cite{xu2024pointllm} & 50.7 & - & - & - & - & - \\
PointLLM~\cite{xu2024pointllm} w/ MI & - & 7.4 & 31.2 & 2.1 & 8.7 & 27.6 \\
\midrule
\rowcolor[gray]{0.7}\textbf{Ours w/o PIT block} & 53.4 & 53.9 & \textbf{52.1} & \textbf{52.9} & \textbf{51.7} & 50.2 \\
\rowcolor[gray]{0.7}\textbf{Ours}  & \textbf{54.2} & \textbf{54.7} & 51.1 & \textbf52.5 & 51.2 & \textbf{51.5} \\
\bottomrule
\end{tabular}
\end{table}

\section{Experiments}
\label{sec:experiments}
We first describe the experimental setup, including datasets, evaluation metrics, baseline models, and shared implementation details. We then present results on our main multi-object comparison task, MO3D, where we compare our Multi-3DLLM against baselines. We evaluate performance on two specialized mini-applications, SM and CC, and then assess foundational capabilities on a standard zero-shot classification benchmark to measure knowledge transfer and potential catastrophic forgetting. Finally, we present ablation studies analyzing key design choices.

\noindent{\textbf{Datasets.}}
Our experiments use MO3D dataset, our primary multi-object comparison dataset, and two application-driven benchmarks, SM (geometric compatibility) and CC (edit-grounded understanding). For the zero-shot classification task, we also use the ModelNet40 benchmark~\cite{wu20153d}.

\noindent{\textbf{Evaluation Metrics.}}
To comprehensively assess performance, we use task-specific metrics. For binary questions, including MO3D (holistic) and CC (verify), we adopt a two-stage protocol. First, Binary Accuracy (B) checks whether the answer starts with the correct ``Yes'' or ``No'' prefix. For SM, Selection Accuracy (S) measures accuracy in a 4-way choice among ``(1,2)'', ``(1,3)'', ``(2,3)'', and ``None''. Second, for all choice-based tasks (measured by B or S), Reasoning Accuracy (R) is computed on the subset of correctly answered samples: an LLM evaluator (\texttt{gpt-4o-mini}) assigns a binary score. To validate this metric, a human audit of 300 responses showed 91.0\% unanimous agreement with the LLM's judgments. For open-ended questions, we report Semantic Accuracy (M), where the evaluator assigns a correctness score using task-specific criteria.

For MO3D, it checks semantic alignment with the ground-truth text and, if misaligned, then checks whether the answer is visually grounded in the multi-view images; a response is marked incorrect only if both tests fail. For CC (delta caption), we decompose the ground truth into edit components (e.g., ``thicken chair legs'', ``soften edges'') and score answers on a 10-point scale based on how many components are correctly identified, missed, or contradicted, with any contradiction yielding a score of 0. For zero-shot classification on ModelNet40~\cite{wu20153d}, we use CLIP-based~\cite{radford2021learning} retrieval accuracy, comparing the cosine similarity between the embedding of the generated text and those of the 40 class names.

\noindent{\textbf{Baseline Models.}}
We compare our approach against a representative set of state-of-the-art object-centric 3D large language models (3D-LLMs), including the original PointLLM~\cite{xu2024pointllm}, ShapeLLM~\cite{qi2024shapellm}, and MiniGPT-3D~\cite{tang2024minigpt3d}. All selected 3D baselines are conversational and accept point clouds as their primary 3D input. In addition, we compare against leading 2D Vision-Language Models (VLMs), specifically LLaVA~\cite{liu2023visual} and Molmo~\cite{deitke2025molom}.

\noindent{\textbf{Input Format for Baselines.}}
For fairness, we match each baseline with an input format suited to its modality. 3D-LLMs, originally single-input, are extended to our multi-object setting via a point cloud concatenation strategy that merges multiple point clouds into one 8,192-point cloud while preserving geometric separation (see Supplementary Material). For 2D-VLMs, we replace point cloud inputs with rendered views of the point clouds, using either one view (1-view) or two opposite views (2-view) per object. Additional baseline comparisons (including scene-level and text-only models) and zero-shot evaluations on real-world datasets for Shape Mating are detailed in the Supplementary Material.

\noindent{\textbf{Implementation Details.}}
Our Multi-3DLLM extends PointLLM, reusing the Point-BERT encoder and Vicuna-7B backbone while adding key architectural changes for multi-object comparison (see Section \ref{sec:model_architecture}). All experiments were conducted on 8 NVIDIA H200 GPUs; remaining training details are given in the supplementary.

\begin{table*}[t]
\centering
\caption{
    Ablation study on the training data mixture. We compare our final model (trained on a uniform mixture of all three datasets) against specialized models trained \textit{only} on a single downstream task. This analysis measures the effect of co-training with our main \textbf{MO3D} dataset on the performance of the mini-applications.
}
\vspace{-0pt}
\label{tab:ablation_training_mixture}
\resizebox{\textwidth}{!}{%
\begin{tabular}{l | c | c | cc | cc | cc | c}
\toprule
\multirow{4}{*}{\textbf{Training Strategy}} 
& \multicolumn{4}{c|}{\textbf{MO3D}} 
& \multicolumn{2}{c|}{\textbf{Shape Mating}} 
& \multicolumn{3}{c}{\textbf{Change Captioning}} \\ 
\cmidrule(lr){2-5} \cmidrule(lr){6-7} \cmidrule(lr){8-10} 
& \multicolumn{1}{c|}{\textbf{Positional (\%)}} 
& \multicolumn{1}{c|}{\textbf{Comparative (\%)}} 
& \multicolumn{2}{c|}{\textbf{Holistic (\%)}} 
& \multicolumn{2}{c|}{\textbf{Selection (\%)}} 
& \multicolumn{2}{c|}{\textbf{Verify (\%)}}   
& \multicolumn{1}{c}{\textbf{Delta Caption (\%)}} \\

& M & M & B & R & S & R & B & R & M \\
\midrule

Multi-3DLLM (MO3D-only)
& 45.5 & 21.9 & \textbf{84.6} & 53.0 & - & - & - & - & - \\

Multi-3DLLM (ShapeMating-only) 
& - & - & - & - & 34.4 & 36.7 & - & - & - \\
 
Multi-3DLLM (ChangeCaptioning-only) 
& - & - & - & - & - & - & 49.7 & 34.0 & 48.2 \\
 
\midrule
\rowcolor[gray]{0.7}\textbf{Multi-3DLLM (All Tasks)} 
& \textbf{56.3} & \textbf{33.8} & 81.7 & \textbf{57.2} & \textbf{37.1} & \textbf{36.8} & \textbf{51.2} & \textbf{37.3} & \textbf{51.0} \\
 
\bottomrule
\end{tabular}
}
\vspace{-10pt}
\end{table*}

\subsection{Multi-Object Comparison (MO3D)}
\label{ssec:mo3d_results}

\noindent{\textbf{Objective.}}
We evaluate Multi-3DLLM on the MO3D dataset to assess its multi-object comparison capabilities against SOTA 3D-LLM and 2D-VLM baselines.

\noindent{\textbf{Experimental Design.}}
We test all models on the MO3D test set. Our model is compared with 3D-LLMs (via point cloud concatenation) and 2D-VLMs (via rendered images). We report Semantic Accuracy (M) for positional and comparative tasks, and Binary (B) / Reasoning (R) Accuracy for the holistic task.

\noindent{\textbf{Results.}}
Table~\ref{tab:pilot-results} shows that our Multi-3DLLM consistently outperforms all baselines across the four MO3D metrics. It scores 56.3 on Positional (M) and 81.7 on Holistic (B), substantially higher than the best 3D-LLM baseline, ShapeLLM, which scored 22.6 and 49.8, respectively. Our model also exceeds the strongest 2D-VLM baselines, LLaVA and Molmo.
The gap is most pronounced on open-ended tasks: in Comparative (M), our model achieves 33.8, nearly three times the next best 3D baseline, ShapeLLM, which scored 11.2, and the best 2D baseline, LLaVA, which scored 11.7, demonstrating a significantly stronger ability to articulate fine-grained differences.
A key observation from Table~\ref{tab:pilot-results} is that both 2D-VLM and 3D-LLM baselines fail on the open-ended Positional and Comparative tasks, even though 3D-LLMs see full 3D point clouds and 2D-VLMs see 2D projections. This suggests that while captions offer a strong prior, 2D-VLMs struggle with geometric awareness, and existing 3D-LLMs face challenges in multi-object comparison.

\subsection{Performance on Mini-Applications}
\label{ssec:miniapp_results}

\noindent{\textbf{Objective.}}
Beyond the main comparison dataset, we evaluate two specialized mini-applications, SM and CC, to test whether our model, which is trained on a holistic mixture of tasks, supports fine-grained geometric compatibility and edit-grounded instruction following, and to assess the value of geometry-aware 3D modeling. Both tasks heavily depend on detailed 3D geometry. We therefore hypothesize that 2D-VLM baselines, which only process 2D projections and lack geometric awareness, will exhibit clear limitations, underscoring the value of 3D-native models for complex spatial reasoning.

\noindent{\textbf{Experimental Design.}}
We evaluate our model, Multi-3DLLM, on the held-out test sets for both SM and CC. We compare its performance against the full suite of 3D-LLM and 2D-VLM baselines to test our hypothesis on the necessity of 3D-native models. Performance is measured using the task-specific metrics defined in our Evaluation Metrics section: Selection (S) and Reasoning (R) Accuracy for SM, Binary (B) and Reasoning (R) Accuracy for CC (Verify), and Semantic Accuracy (M) for CC (Delta Caption).

\noindent{\textbf{Results.}}
Table~\ref{tab:pilot-results} summarizes results on the two mini-applications and provides evidence for the necessity of geometrically-aware 3D-LLMs. On SM, our hypothesis from the Objective is strongly validated: all 2D-VLM and 3D-LLM baselines score well below the 25\% four-way chance level, indicating a failure to capture the geometric compatibility signal. In contrast, Multi-3DLLM achieves a Selection (S) score of 37.1\%, outperforming all baselines. Moreover, all baseline models score near-zero on the Reasoning (R) metric, showing that their already-poor selections lack valid geometric justification and are effectively random. CC yields a more nuanced picture. For the generative Delta Caption (M) task, we again see near-zero performance from all 2D and 3D baselines, while our model reaches 51.0\% and is the only architecture that demonstrates comprehension. The Verification (B) task is exceptionally challenging, with all models performing near the 50.0\% chance baseline (further analysis on real-world domain transfer and enhanced reasoning strategies is provided in Supp).

\subsection{Evaluation on Zero-Shot Classification}
\label{ssec:zeroshot_results}

\noindent{\textbf{Objective.}}
We evaluate the model's foundational capabilities using a zero-shot classification task. This evaluation serves two main purposes.
We first test whether the model can ground positional language (\eg, ``the first object'') to the correct point cloud in multi-object scenes, a critical skill for all our tasks. We then examine how our holistic training strategy affects this fundamental skill. By comparing our final model with the original PointLLM, we assess whether this training strategy leads to positive knowledge transfer or catastrophic forgetting, while comparison with the variant \textit{w/o PIT block} isolates the impact of our architectural changes on high-level semantic recognition.

\noindent{\textbf{Experimental Design.}}
We perform zero-shot classification on ModelNet40 benchmark~\cite{wu20153d} under three settings (see Table~\ref{tab:mo3d_adapt_results}): (1) Single-Object, (2) 2-Input, and (3) 3-Input. In the multi-object settings, the model receives a set of point clouds from different ModelNet40 classes, with one designated as the target and the rest as distractors, and is queried with an ordinal prompt such as ``What is the class of the first object?'' or ``Identify the category of the n-th object'' For the 2-Input task we query both positions ($M{=}1$, $M{=}2$); for the 3-Input task we query all three ($M{=}1$, $M{=}2$, $M{=}3$). Classification accuracy is measured using the CLIP-based~\cite{radford2021learning} retrieval protocol from the original PointLLM.

\noindent{\textbf{Results.}}
Table~\ref{tab:mo3d_adapt_results} shows that \textbf{PointLLM w/ MI}, which only adds multi-input capacity without multi-object tuning, fails critically: it exhibits strong positional bias~\cite{zheng2023judging}, with accuracy collapsing on early objects (\eg, a score of 2.1 for $M{=}1$ in the 3-Input task) and recovering only for the final object. This indicates that positional grounding must be learned. In contrast, fine-tuned models achieve stable accuracy across all positional queries, avoid catastrophic forgetting, and even surpass PointLLM on the single-object setting. This shows that task-mixture training yields positive knowledge transfer and strengthens semantic understanding. Finally, our full model performs similarly to the Multi-3DLLM (\textit{w/o PIT}) variant, suggesting that the PIT block, while crucial for fine-grained comparative and geometric tasks (Sec.~\ref{ssec:mo3d_results}), is not required for high-level semantic classification and does not degrade its basic capability.

\begin{table*}[t]
\centering
\caption{
Quantitative comparison of attention patterns between our Multi-3DLLM model and a model trained only on MO3D baseline.
Metrics are computed on 266 object instances from the MO3D test set. Sparsity (Gini) measures attention inequality (1 = max concentration). Entropy measures attention uniformity (0 = max concentration). Top-K\% Conc indicates proportion of attention mass on the top K\% of points. Effective Rank approximates the number of points receiving substantial attention, computed as $\exp(H)$ where $H$ is the entropy.
}
\label{tab:attention_analysis}
\small
\begin{tabular}{lccccc}
\toprule
\textbf{Model} & \textbf{Sparsity (Gini)} & \textbf{Entropy} & \textbf{Top-10\% Conc.} & \textbf{Top-20\% Conc.} & \textbf{Effective Rank}  \\
\midrule
Multi-3DLLM (MO3D-Only) & 0.605 & 0.822 & 0.538 & 0.652 & 2209 \\
\rowcolor[gray]{0.7}\textbf{Multi-3DLLM} & \textbf{0.726} & \textbf{0.716} & \textbf{0.675} & \textbf{0.763} & \textbf{1088} \\
\bottomrule
\end{tabular}
\end{table*}

\subsection{Ablation Studies}

\noindent\textbf{Impact of the PIT block.}
We ablate the PIT block by comparing three architectural variants, shown in Table~\ref{tab:pilot-results}. \emph{Multi-3DLLM (w/o PIT block)}, hereafter \emph{No-Interaction}, feeds the concatenated tokens to the LLM without cross-object mixing. \emph{Multi-3DLLM}, hereafter \emph{Object-Level}, uses an object-level encoder that mean-pools each object's patch tokens, runs a self-attention over the resulting $N$ object tokens, and broadcasts one residual update per object back to its $T$ patch tokens, as in prior scene-level designs~\cite{wang2023chat3d}. Our full model, \emph{Multi-3DLLM (Ours)}, hereafter \emph{Ours (PIT)}, instead applies a \emph{patch-level} Transformer that computes per-token residuals, as detailed in Sec.~\ref{sec:model_architecture}.

The results in Table~\ref{tab:pilot-results} validate our patch-level interaction design. On MO3D, \emph{Ours (PIT)} improves fine-grained positional and comparative reasoning over \emph{No-Interaction} by 10.8 points on Positional (M) and 12.0 points on Comparative (M), showing that patch-level cross-object mixing is crucial for relational understanding. \emph{Object-Level} performs particularly poorly on SM: its Selection accuracy drops to 25.0 versus 34.4 for \emph{No-Interaction}, and its Rationale accuracy to 23.7 versus 36.7. This indicates that mean-pooling and broadcasting a single object-wise residual washes out token-level cues such as mating interfaces. By computing token-specific updates, \emph{Ours (PIT)} preserves these local correspondences and achieves the best SM scores, 37.1 on Selection and 36.8 on Rationale. On CC, Ours (PIT) also performs strongly suggesting that patch-level interactions support more specific and faithful descriptions. Overall, \emph{Ours (PIT)} yields the most robust gains on tasks requiring both inter-object and intra-object (patch-level) reasoning.


\noindent\textbf{Attention analysis}
\label{ssec:ablation_attention}
We evaluate our training strategy by comparing the attention patterns of a model trained on MO3D + Mini-apps with a MO3D-only baseline. We analyze 266 object-level attention distributions from 100 MO3D test sets and compute four metrics (Table~\ref{tab:attention_analysis}): Gini Coefficient, Normalized Entropy, Top‑K Concentration, and Effective Rank. Effective Rank, defined as $\exp(H)$ where $H$ is the Shannon entropy~\cite{shannon1948mathematical}, approximates the number of points receiving substantial attention, which we call the “used points”. Table~\ref{tab:attention_analysis} shows that multitask learning yields markedly more efficient and sparser attention: the Effective Rank drops by 50.7\%, indicating that our model processes less than half as many points while achieving higher overall performance (Table~\ref{tab:ablation_training_mixture}). We hypothesize that this acts as an attention regularizer---by solving fine-grained tasks such as SM, the model learns to aggressively discard irrelevant information and focus on salient geometric features (Fig.~\ref{fig:attention_vis}). The resulting sparsity not only improves performance but also offers promising opportunities for future computational optimizations and enhanced interpretability.

\begin{figure}[t]
\centering
\includegraphics[width=\linewidth]{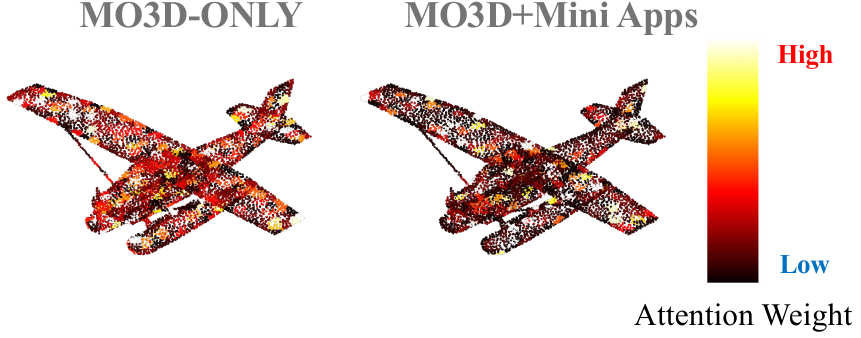} 
\caption{
    \textbf{Attention visualization comparison.} 
    Attention weights overlaid on a 3D point cloud from the MO3D test set. 
    Left (MO3D Only): The task-specific model shows distributed, diffuse attention across the entire object.
    Right (MO3D + Mini-apps): Our multitask model exhibits concentrated and sparse attention.
    Attention weights are mapped to colors, with yellow indicating high attention and darker colors indicating low attention. The multitask model achieves competitive performance while attending to significantly fewer points (see Table~\ref{tab:attention_analysis}).
}
\label{fig:attention_vis}
\vspace{-10pt} 
\end{figure}

\section{Conclusion}
We expanded conventional 3D-LLMs, previously limited to single object, to handle multi-object point clouds for comparison, shape mating, and change captioning. The MO3D dataset provides a robust benchmark, while our Multi-3DLLM, trained on a holistic data mixture with a minimal patch-level adaptation, significantly outperforms prior methods (e.g., PointLLM). We further demonstrate its practical utility on complex application-driven tasks.

\noindent\textbf{Limitations and future work.}
Our pipeline relies on existing captions, inheriting potential biases in description granularity. Extending to scenes with more objects introduces token budget trade-offs, and transferring to real-world scans highlights point density challenges. Future work will explore hierarchical part-based annotations, efficient token scaling strategies, and robust real-world domain adaptation.


{
    \small
    \bibliographystyle{ieeenat_fullname}
    \bibliography{main}
}

\clearpage
\appendix
\section*{Supplementary Material}

\section{Implementation Details}
\subsection*{Model Checkpoints}
We initialized our system using publicly available pre-trained models from the PointLLM~\cite{xu2024pointllm} framework:

\noindent{\textbf{Language Model:}} Vicuna-7B-v1.5 (lmsys/vicuna-7b-v1.5)~\cite{chiang2023vicuna}

\noindent{\textbf{Point Cloud Encoder:}} Point-BERT~\cite{yu2022pointbert}, pre-trained via ULIP2~\cite{xue2024ulip2}

\noindent{\textbf{CLIP used for classification, grouping:}} OpenCLIP ViT-L/14 backbone~\cite{radford2021learning}

\subsection{Training Hyperparameters}
We detail the hyperparameters used for our two-stage training strategy in Table~\ref{tab:hyperparams}. All models were trained using the AdamW optimizer with a cosine learning rate scheduler and a warmup ratio of 0.03.

\begin{table}[h]
\centering
\caption{\textbf{Training Hyperparameters.}}
\label{tab:hyperparams}
\resizebox{1.0\linewidth}{!}{%
\begin{tabular}{l|cc}
\toprule
\textbf{Parameter} & \textbf{Phase 1} & \textbf{Phase 2} \\
& Feature Alignment & Holistic Task-Mixture \\
\midrule
Batch Size & 16 & 14 \\
Learning Rate & 2e-3 & 2e-5 \\
Weight Decay & 0.0 & 0.0 \\
Number of Epochs & 3 & 3 \\
LR Scheduler & Cosine w/ Warmup & Cosine w/ Warmup \\
Warmup Ratio & 0.03 & 0.03 \\
\midrule
\textbf{Trained Params} & Projector & Projector, PIT block, LLM \\
\midrule
\textbf{Frozen} & Point Encoder & Point Encoder \\
\textbf{Params} & PIT block, LLM & \\
\bottomrule
\end{tabular}
}
\end{table}

\subsection*{Hardware}
Training was conducted on 8 NVIDIA H200 GPUs (140GB VRAM). 

\noindent{\textbf{Training Time.}}
The training process was completed in two phases with the following durations: 70 minutes for feature alignment (Phase 1), 12 hours for Training on holistic mixture of datasets (MO3D, Shape Mating, Change Captioning).

\section{Dataset Generation Details}
\subsection{MO3D Dataset}
\noindent\textbf{Data-Driven Category Definition.}
To ensure our benchmark covers diverse aspects of 3D objects, we established six core categories through a two-stage process. First, we prompted \textit{Qwen2-72B-Instruct}~\cite{bai2023qwen} to identify common attribute types from 70k samples of Objaverse-Cap3D~\cite{deitke2023objaverse, luo2023scalable} captions. Then, we manually grouped these outputs into six semantically distinct and comprehensive categories, followed by manual curation. These categories, detailed in Table~\ref{tab:category_curation}, form the foundation for our balanced generation pipeline.

\begin{figure*}[t]
    \centering
    \includegraphics[width=0.99\textwidth]{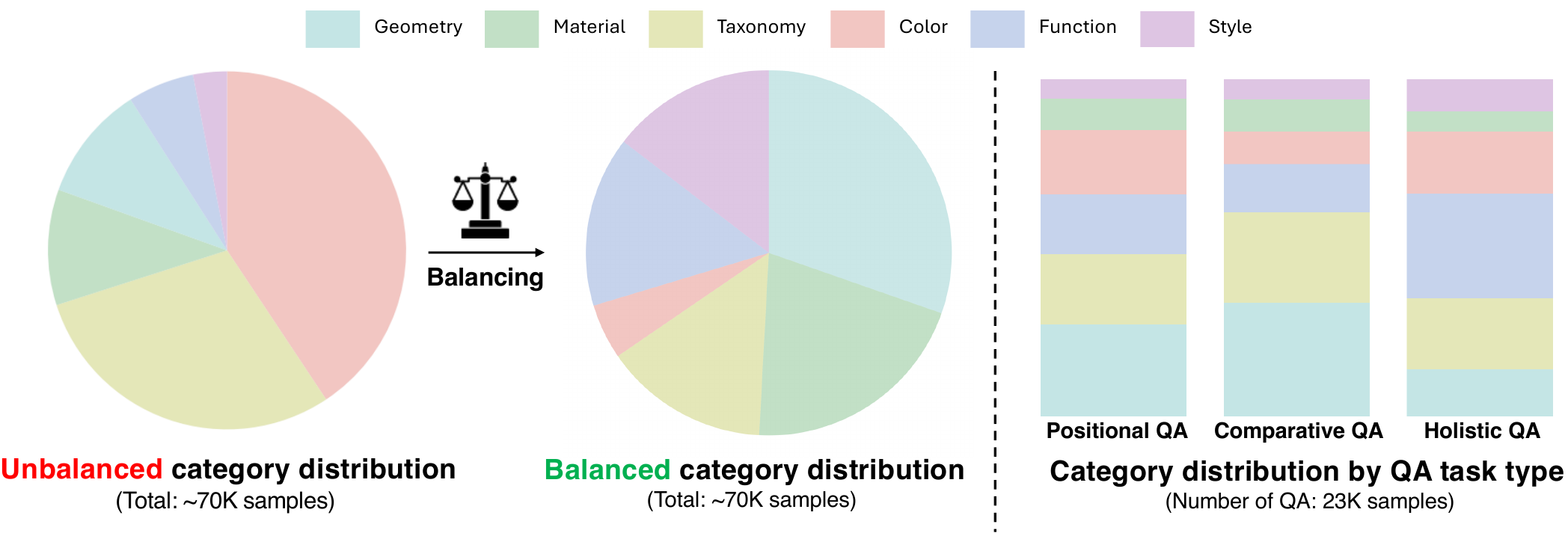}
    \caption{
    \textbf{Category Distribution and Weighted Balancing.} 
    We transform the naturally occurring, unbalanced attribute distribution (left) into a curated target distribution (middle). 
    This process explicitly prioritizes geometric and structural categories (\eg, Geometry, Material) while suppressing superficial visual cues (\eg, Color) to enforce 3D understanding. 
    The stacked bars (right) verify that this balanced distribution is consistently applied across all three task types: Positional, Comparative, and Holistic QA.
}
    \label{fig:category_distribution}
    \vspace{-10pt}
\end{figure*}

\begin{table}[h]
\centering
\caption{
    \textbf{Overall Statistics of the MO3D Dataset.} 
    The dataset maintains a perfectly balanced distribution across the three core tasks and a near-even split between positive and negative object groupings.
}
\label{tab:mo3d_stats}
\resizebox{0.95\linewidth}{!}{%
\begin{tabular}{c|ccc|cc}
\toprule
\multirow{2}{*}{\textbf{Total Samples}} & \multicolumn{3}{c|}{\textbf{Task Types}} & \multicolumn{2}{c}{\textbf{Group Types}} \\
& \textbf{Positional} & \textbf{Comparative} & \textbf{Holistic} & \textbf{Positive} & \textbf{Negative} \\
\midrule
\textbf{69,996} & 23,332 & 23,332 & 23,332 & 34,926 & 35,070 \\
(100\%) & (33.3\%) & (33.3\%) & (33.3\%) & (49.9\%) & (50.1\%) \\
\bottomrule
\end{tabular}
}
\end{table}

\begin{table}[h]
\centering
\caption{
    \textbf{Distribution of Question Categories.} We compare our design target weights with the actual distribution in the final dataset. 
    The resulting distribution closely matches our goal of prioritizing geometric and structural comparison.
}
\label{tab:mo3d_categories}
\small
\begin{tabular}{l cc}
\toprule
\textbf{Category} & \textbf{Target Weight} & \textbf{Actual Count (\%)} \\
\midrule
Geometry / Structure & 30\% & 21,264 (30.4\%) \\
Material             & 20\% & 14,316 (20.5\%) \\
Function             & 15\% & 10,578 (15.1\%) \\
Taxonomy             & 15\% & 10,200 (14.6\%) \\
Style / Aesthetics   & 15\% & 10,206 (14.6\%) \\
Color                & 5\%  & \phantom{0}3,432 \phantom{0}(4.9\%) \\
\bottomrule
\end{tabular}
\end{table}

\begin{table*}[t]
\centering
\caption{
The six core reasoning categories curated from our data-driven analysis. We define each category and provide examples of the raw, LLM-discovered themes that were merged into it.}
\label{tab:category_curation}
\resizebox{\textwidth}{!}{%
\begin{tabular}{p{3.2cm} | p{6.3cm} | p{6.3cm}}
\toprule
\textbf{Core Category} & \textbf{Description and Example Question Focus} & \textbf{Examples of Merged Raw Categories} \\
\midrule
\textbf{geometry/structure} & The shape, size, number of parts, and structural complexity of an object. \newline \textit{e.g., ``Does it have more than four legs?''} & \texttt{shape}, \texttt{size}, \texttt{components}, \texttt{part}, \texttt{feature}, \texttt{features}, \texttt{detail}, \texttt{base}, \texttt{form} \\
\midrule
\textbf{taxonomy} & The general class, type, or identity of an object. \newline \textit{e.g., ``Is this object a piece of furniture?''} & \texttt{objecttype}, \texttt{type}, \texttt{object\_type}, \texttt{object}, \texttt{theme}, \texttt{category} \\
\midrule
\textbf{function} & The intended purpose, use, or action associated with an object. \newline \textit{e.g., ``What is this object used for?''} & \texttt{function}, \texttt{action}, \texttt{purpose}, \texttt{usage}, \texttt{activity} \\
\midrule
\textbf{material} & The physical substance an object is made of. This is distinct from color. \newline \textit{e.g., ``Is the frame made of metal or wood?''} & \texttt{material}, \texttt{substance}, \texttt{composition} \\
\midrule
\textbf{style/aesthetics} & The visual style, era, pattern, decoration, or contextual setting of an object. \newline \textit{e.g., ``Does this object have a writing on the surface?''} & \texttt{style}, \texttt{decoration}, \texttt{pattern}, \texttt{design}, \texttt{era}, \texttt{location}, \texttt{context} \\
\midrule
\textbf{color} & The surface color or hue of an object or its parts. \newline \textit{e.g., ``Is the main color of the object brown?''} & \texttt{color}, \texttt{hue}, \texttt{finish} \\
\bottomrule
\end{tabular}%
}
\end{table*}
\label{sec:appendix}

\noindent\textbf{Group Formation Strategy.}
The foundation of each query is a meaningful group of objects. We formulated these groups based on the semantic similarity of captions from the Objaverse-Cap3D corpus. To quantify similarity efficiently at scale, we computed embeddings using a pre-trained CLIP text encoder (ViT-L/14)~\cite{radford2021learning} and leveraged Faiss~\cite{johnson2019billion} to pre-compute and cache the top-50 nearest neighbors for every object.

At generation time, we select an anchor object and $n-1$ companions. We specifically focus on group sizes of $n \in \{2, 3\}$. The inclusion of 3-object groups is a deliberate design choice: a group of three is the minimal configuration required to introduce complex relational concepts, such as identifying a semantic ``outlier'' or finding a ``majority'' property, which are impossible with only pairwise comparisons.
To ensure a balanced distribution of comparison scenarios, we employ a Dual Sampling strategy with specific thresholds:
\begin{itemize}
    \item Positive Sampling: Selects companion objects from the anchor's top-50 semantically similar neighbors. This results in groups with high conceptual overlap, suitable for fine-grained comparison (e.g., distinguishing between two different chairs).
    \item Negative Sampling: Selects companions from the pool of items that do not appear in the anchor's top-50 neighbors. This produces groups with low conceptual overlap for broader distinctions.
\end{itemize}
Finally, the sequence of selected objects is randomly shuffled to ensure the task is order-invariant and to prevent the model from learning positional biases.

\noindent\textbf{Guided Question Generation.}
We use GPT-4 to generate QA pairs, inputting both the Cap3D captions and multi-view renderings. To prevent hallucinations and ensure quality, we enforce a strict Prompt Hierarchy:
\begin{itemize}
    \item Grounding Constraint (Highest Priority): All information must be strictly visually grounded in the provided inputs. Invented details are prohibited.
    \item Task-Specific Constraints : Each task type (e.g., geometry) enforces specific keywords (e.g., use ``shape'' instead of ambiguous ``feature'').
    \item Category Coverage: We use weighted balancing to target specific categories, prioritizing geometric understanding over simpler visual cues as shown in Figure~\ref{fig:category_distribution}. If a category is not applicable (e.g., no color information), the model falls back to a valid alternative.
\end{itemize}
The full prompt used for generation is provided in Figure~\ref{fig:prompt_mo3d}.

\noindent\textbf{Balance Correction.}
To mitigate linguistic priors (\eg, the tendency to answer "Yes"), we implement a post-hoc balance correction. We analyze the \texttt{holistic} subset and identify any imbalance. We then regenerate a subset of samples using a constrained prompt that forces the generation of a "No" question (e.g., asking about a property \textit{not} shared by the group), ensuring a near 50:50 distribution in the final dataset.

\noindent\textbf{Dataset Statistics.}
The final MO3D dataset consists of approximately 70k multi-object instruction-following examples. As detailed in Table~\ref{tab:mo3d_stats}, the dataset is perfectly balanced across the three core tasks (Positional, Comparative, Holistic), with each constituting exactly one-third of the data. The object grouping strategies (Positive vs. Negative sampling) are also balanced near 50:50 to ensure diverse comparison scenarios. Furthermore, Table~\ref{tab:mo3d_categories} demonstrates that our weighted sampling strategy successfully aligned the generated question categories with our target distribution, prioritizing geometric and structural understanding (30.4\%) over simpler attributes like color (4.9\%). We further analyze the linguistic complexity in Table~\ref{tab:length_stats} and Figure~\ref{fig:mo3d_length_dist}. 

\noindent\textbf{Data Reliability and Intrinsic Ambiguity.}
To rigorously validate the reliability of the MO3D dataset, we conducted a human audit on 500 randomly sampled QA pairs from the test set. Three independent annotators evaluated each pair, achieving a strong unanimous agreement rate of 81.0\%. 
Through this audit, we found that the remaining disagreements did not stem from incorrect ground truths, but rather from the \emph{intrinsic ambiguity} of 3D object interpretation. As illustrated in Figure~\ref{fig:intrinsic_ambiguity}, subjective perceptions of highly abstract shapes or partial occlusions naturally lead to divergent valid interpretations among humans. This highlights the inherent complexity of 3D relational tasks compared to standard 2D QA. Crucially, when evaluated on this high-quality unanimous subset, our model achieves an average accuracy of 59.2\% across the MO3D tasks, closely aligning with our 57.3\% average on the full test set. This verifies that intrinsic ambiguities do not artificially distort the evaluation metrics.

\begin{figure}[h]
  \centering
  \includegraphics[width=0.98\linewidth, height=3.0cm, keepaspectratio]{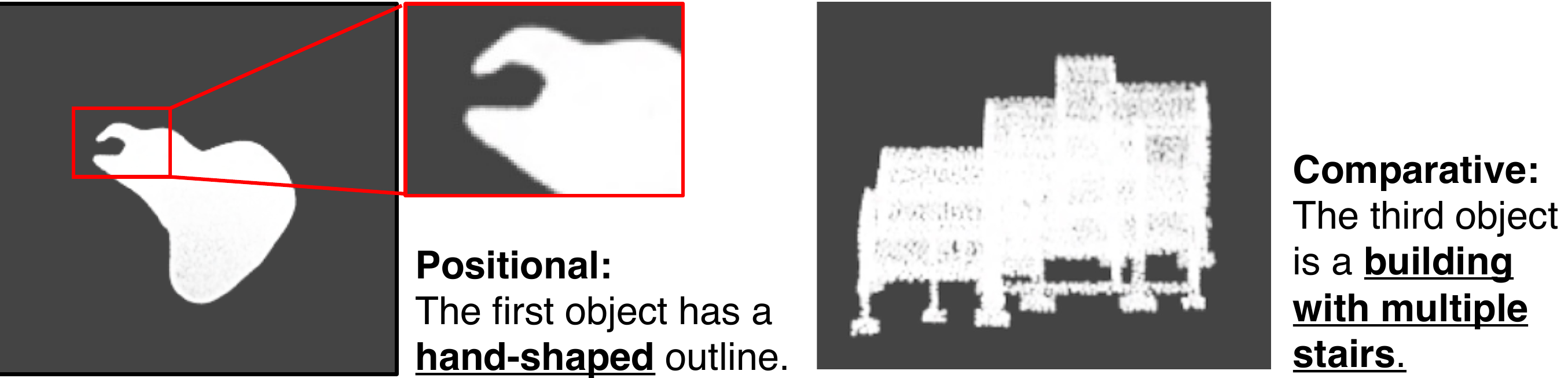}
  \caption{\footnotesize \textbf{Intrinsic Ambiguity.} Disagreements in human verification often arise from subjective perception (\eg abstract shapes or occlusions) }
  \label{fig:intrinsic_ambiguity}
\end{figure}

\begin{table}[h]
\centering
\small 
\setlength{\tabcolsep}{5pt}
\caption{
    \textbf{Linguistic Complexity Statistics.} 
    We report the average and maximum length in words for instructions and responses. 
    Shape Mating features long responses due to the requirement of geometric rationales.
}
\label{tab:length_stats}
\begin{tabular}{l l | cc | cc}
\toprule
\multirow{2}{*}{\textbf{Dataset}} & \multirow{2}{*}{\textbf{Task Type}} & \multicolumn{2}{c|}{\textbf{Instruction}} & \multicolumn{2}{c}{\textbf{Response}} \\
& & \textbf{Mean} & \textbf{Max} & \textbf{Mean} & \textbf{Max} \\
\midrule
\multirow{3}{*}{MO3D} 
& Positional & 10.8 & 21 & 10.0 & 25 \\
& Comparative & 13.3 & 30 & 18.1 & 33 \\
& Holistic & 9.9 & 18 & 12.8 & 31 \\
\midrule
Shape Mating & Selection & 31.8 & 34 & 73.5 & 130 \\
\midrule
\multirow{2}{*}{Change Cap.} 
& Verify & 64.3 & 191 & 25.4 & 83 \\
& Delta & 16.8 & 18 & 25.6 & 96 \\
\bottomrule
\end{tabular}
\end{table}

\subsection{Mini-App A: Shape Mating Details}

\noindent\textbf{Data Construction and Sampling.}
We source base 3D meshes from the Thingi10K dataset~\cite{zhou2016thingi10k} and generate mating pairs using the \texttt{cut shell} operation from Neural Shape Mating~\cite{chen2022neural}. We utilize five distinct cut geometries: \textit{Planar}, \textit{Sine}, \textit{Square}, \textit{Pulse}, and \textit{Parabolic}. Crucially, we selected the \texttt{cut shell} operation to ensure domain consistency with our point cloud encoder, Point-BERT~\cite{yu2022pointbert}. Since Point-BERT is pre-trained solely on object surface points rather than interior cross-sections, the introduction of artificial flat cut-faces inherent to solid cutting methods would result in a domain gap. Consequently, we utilize the \texttt{cut shell} method to preserve the surface-shell characteristic, thereby aligning the input distribution with the pre-training regime of the encoder. Finally, we perform uniform random sampling on the mesh surface to generate dense point clouds of 8,192 points per part (Figure~\ref{fig:sm_generation}), capturing the fine-grained geometric details required for the mating task.

To strictly enforce the 4-choice classification task, we employ a targeted sampling logic for scene composition. For \textit{1-Mate (Positive)} scenarios, we randomly select a valid ground-truth pair (Part A and Part B from the same instance) and sample a third "decoy" part. This decoy is carefully selected to be non-mating due to specific reasons, such as originating from a different object, a different cut type, or a different cut position (Phase Mismatch), and its position is randomized. For \textit{0-Mate (Negative)} scenarios, we sample three parts such that no combination forms a valid pair, acting as a hard negative that forces the model to verify all possible connections.
\begin{figure}[t]
    \centering
\includegraphics[width=1.0\linewidth]{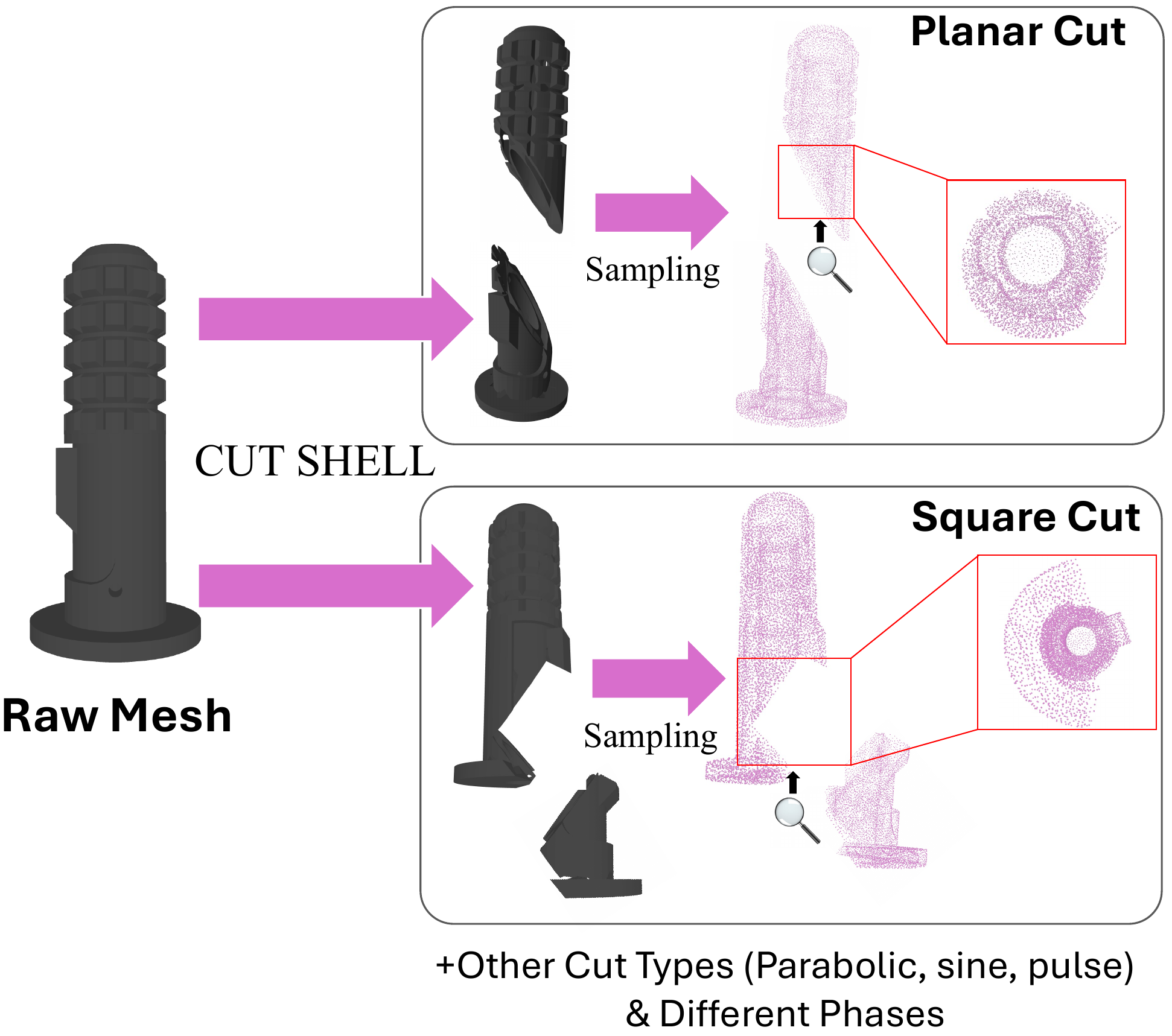}
    \caption{\textbf{Point cloud generation pipeline for Shape Mating.} 
    Starting from a raw mesh (Left), we apply the \texttt{cut shell} operation to split the object into two complementary halves (Middle). 
    Unlike solid cuts, this operation preserves the hollow, surface-only structure of the object. 
    Finally, each part is uniformly sampled into 8,192 points (Right) to serve as the input for our model. 
    }
    \label{fig:sm_generation}
\end{figure}

\begin{figure}[t]
    \centering
    \includegraphics[width=1.0\linewidth]{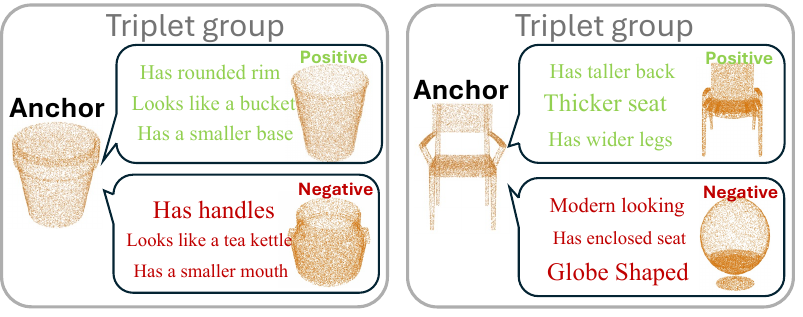}
    \caption{
        \textbf{Hard Negative Sampling for Change Captioning.} 
        We construct contrastive triplets where the \textit{Negative} candidate (red captions) is not random, but a ``hard negative'' sharing the same source \textit{Anchor} as the \textit{Positive} target (green captions). 
        This Negative sample is crucial for the Verification task to generate challenging ``No'' instances
    }
    \label{fig:cc_triplet}
\end{figure}

\noindent\textbf{Question and Rationale Formulation.}
We construct the QA pairs using a two-step process to ensure both linguistic diversity and geometric grounding. First, for the question component, we employ a set of 15 distinct templates (detailed in Table~\ref{tab:sm_templates}) to ensure consistent task formulation while providing linguistic variety. These templates explicitly list the four options and mandate a reasoning-based response. Second, for the rationale component, we train the model to explain why a pair does not mate by automatically assigning structured error tags to non-mating pairs. We then use \textit{GPT-4o-mini} to paraphrase these tags into natural language justifications. The full prompt used for this paraphrasing is provided in Figure~\ref{fig:prompt_sm}. The error types are defined as follows:
\begin{itemize}
    \item \emph{cut mismatch} indicates that the two parts possess disparate cut interfaces, such as a planar surface versus a sinusoidal one.
    \item \emph{object mismatch} signifies that although the parts share the same cut type, they originate from distinct source objects and thus do not align globally.
    \item \emph{phase mismatch} occurs when parts share the same object and cut geometry but are derived from different cut instances or positions, preventing an exact fit.
    \item \emph{same side} denotes topological incompatibility, where the selected parts represent the same side of the object, such as two ``Part A'' components.
\end{itemize}

\noindent\textbf{Dataset Statistics.}
We provide a detailed analysis of linguistic complexity in Table~\ref{tab:length_stats} and visualize the length distributions in Figure~\ref{fig:sm_length_dist}. As shown in the table, Shape Mating involves particularly long responses due to the requirement for detailed geometric reasoning. 

\subsection{Mini-App B: Change Captioning Details}

\noindent\textbf{Data Construction and Sampling.}
We construct this benchmark using the ShapeNet subset of the ShapeTalk dataset~\cite{achlioptas2023shapetalk}. The core unit is a contrastive triplet consisting of an \textit{Anchor}, a \textit{Positive}, and a \textit{Negative} point cloud, paired with an \textit{Instruction}. We employ a strict \emph{Hard Negative Sampling} strategy to ensure difficulty, as illustrated in Figure~\ref{fig:cc_triplet}. For a given Anchor-Positive pair (e.g., "thinner back"), we prioritize sampling a Negative shape that shares the same Anchor but corresponds to a different edit instruction (e.g., "thicker seat"). This forces the model to ground the specific semantic details of the instruction, rather than relying on coarse object recognition. If no such hard negative exists, we fall back to a random object from the same semantic class.

\noindent\textbf{Question and Rationale Formulation.}
We transform the raw triplet data into model inputs using standardized templates. As detailed in Table~\ref{tab:cc_templates}, we employ distinct template sets for each task to ensure consistent definition while introducing phrasing variety. To ensure high-quality linguistic output, we employ \textit{GPT-4o-mini} for both tasks.
For the Verification task, we explicitly randomize the input order of the Anchor and Candidate point clouds to prevent the model from memorizing positional cues (e.g., assuming the answer is always the second object). We used \textit{GPT-4o-mini} to construct a natural language rationale based on the original instruction associated with the Positive/Negative object. For the Delta Captioning task, we similarly use \textit{GPT-4o-mini} to paraphrase and consolidate multiple raw utterances into a single, fluent description.

\noindent\textbf{Dataset Statistics.}
We report the linguistic complexity in Table~\ref{tab:length_stats}. Notably, the Verification task has the longest average instruction length of 64.3 words because the input explicitly includes the full list of geometric requirements (from the instruction) that the model must check.

\subsection{Data Splitting}
\label{ssec:splitting_strategy}

Across all our benchmarks (MO3D dataset, Shape Mating, and Change Captioning), we employ a strict leakage-free splitting strategy to ensure rigorous evaluation, as simple random splitting is insufficient when 3D assets share underlying geometries or appear in multiple grouping scenarios. To prevent data leakage, we utilize a graph-based approach where every unique 3D asset is represented as a node, and edges are drawn between nodes if they appear together in the same sample (\eg, within a triplet) or share the same source object. We then compute the connected components of this graph and atomically assign entire components to a single split (Train or Test). This methodology guarantees that no object instance, nor any of its co-occurring or geometrically related variants, ever leaks across splits, ensuring that our evaluation measures true generalization rather than memorization.

\section{Additional Ablation Studies}
\subsection{Baselines \& Input Format Fairness}
\label{ssec:supp_baselines}
To ensure a fair comparison across fundamentally different architectures, we carefully designed modality-specific input formatting strategies:

\noindent\textbf{Object-Centric 3D-LLMs:} Models like PointLLM and ShapeLLM natively accept a single point cloud. For our multi-object tasks, we implement a \emph{concatenation with separation} strategy. Each object's point cloud is individually normalized into a unit sphere, and then translated along a single axis (e.g., the x-axis) with a fixed margin. This preserves the intrinsic local geometry of each object while keeping them spatially distinct within a single 8,192-point input limit.

\noindent\textbf{Scene-Level Models:} We evaluated scene-level models, including Chat-Scene~\cite{huang2024chat} and LL3DA~\cite{chen2024ll3da}. Since these models are designed to extract objects from a full scene context, we provided them with the explicit centroid coordinates (click) or bounding boxes (BBox) of the target objects, mapping ordinal textual queries (\eg, ``the first object'') to their corresponding spatial prompts. As shown in Table~\ref{tab:combined_res}, despite this explicit localization, these models severely underperformed on MO3D and Shape Mating. This confirms our architectural finding: their object-level token pooling smooths out the fine-grained local geometry required for detailed comparison.

\noindent\textbf{Text-Only Baseline (Language Bias):} To isolate the contribution of 3D geometric reasoning from linguistic priors, we evaluated a text-only baseline using Vicuna-7B (Table~\ref{tab:combined_res}). We provided the model with the ground-truth Cap3D text descriptions instead of visual inputs. The text-only model achieved scores on MO3D of 52.0 (Positional M), 28.9 (Comparative M), 70.5 (Holistic B), and 50.3 (Holistic R). While these scores demonstrate that the ground-truth captions offer a very strong semantic prior, our Multi-3DLLM consistently outperforms this text-only baseline across all metrics (e.g., 56.3 on Positional, 33.8 on Comparative). This confirms that our model's gains stem from genuine 3D geometry processing rather than merely exploiting language biases.

\begin{table}[t]
\centering
\caption{Additional results on MO3D: scene-level and text-only baselines.}
\label{tab:combined_res}
\footnotesize
\setlength{\tabcolsep}{3pt}
\renewcommand{\arraystretch}{0.95}
\begin{tabular}{cl|cccc}
\toprule
\multicolumn{2}{l|}{Model} & Pos. & Comp. & Hol. (B) & Hol. (R) \\
\midrule
\multirow{3}{*}{\textit{Scene-level}} & Chat-Scene & 10.8 & 3.0 & 49.7 & 31.6 \\
& LL3DA (Click) & 17.8 & 4.5 & 48.8 & 26.5 \\
& LL3DA (Bbox) & 18.2 & 2.9 & 48.8 & 25.9 \\
\midrule
\textit{Text-only} & Vicuna-7B (GT captions) & 52.0 & 28.9 & 70.5 & 50.3 \\
\midrule
\multicolumn{2}{l|}{Ours (Multi-3DLLM)} & \textbf{56.3} & \textbf{33.8} & \textbf{81.7} & \textbf{57.2} \\
\bottomrule
\end{tabular}
\end{table}

\subsection{Architectural Ablation: Interaction Mechanics}
\label{ssec:ablation_interaction_extended}

\noindent\textbf{Motivation.}
In the main paper, we demonstrated that the \emph{Object-Level} interaction fails on geometric tasks. To further investigate whether this failure stems from the loss of salient features (due to mean pooling) or the loss of spatial resolution (due to object-wise broadcasting), we evaluate two additional architectural variants.

\noindent\textbf{Variants.}
\begin{itemize}
    \item \textbf{Object-Level (Max Pooling):} Similar to the mean-pooling baseline, this variant aggregates object tokens into a single vector. However, it uses \emph{max-pooling} to capture the most salient features (e.g., sharp corners or handles) across the patch tokens. The updated residual is then broadcast uniformly to all patches of the object. This tests if preserving salient features is sufficient for geometric reasoning.
    
    \item \textbf{Micro-Token Interaction:} This variant operates at an intermediate granularity. Instead of collapsing an object into a single vector, we compress the object's patch tokens into $M=32$ representative \emph{``micro-tokens''} inspired by \emph{0M-Pooling} from ~\cite{tang2025more}. This is achieved by aggregating patch tokens into distinct clusters based on feature similarity, thereby reducing redundancy while preserving diverse local features. The interaction module processes these micro-tokens, and the update is redistributed to the original patches via a cross-attention mechanism, allowing for spatially varying updates.
\end{itemize}

\begin{table*}[t]
\centering
\caption{
    Extended ablation on interaction mechanics. We compare different pooling strategies (Mean vs. Max) and granularities (Object vs. Micro vs. Patch).
    \emph{Micro-Token} uses 32 representative tokens per object. \emph{w/ PIT} uses full patch-level interaction.
}
\label{tab:ablation_interaction_extended}
\resizebox{\textwidth}{!}{%
\begin{tabular}{l | c | c | cc | cc | cc | c}
\toprule
\multirow{4}{*}{\textbf{Interaction Mechanism}} 
& \multicolumn{4}{c|}{\textbf{MO3D}} 
& \multicolumn{2}{c|}{\textbf{Shape Mating}} 
& \multicolumn{3}{c}{\textbf{Change Captioning}} \\ 
\cmidrule(lr){2-5} \cmidrule(lr){6-7} \cmidrule(lr){8-10} 
& \multicolumn{1}{c|}{\textbf{Positional (\%)}} 
& \multicolumn{1}{c|}{\textbf{Comparative (\%)}} 
& \multicolumn{2}{c|}{\textbf{Holistic (\%)}} 
& \multicolumn{2}{c|}{\textbf{Selection (\%)}} 
& \multicolumn{2}{c|}{\textbf{Verify (\%)}}   
& \multicolumn{1}{c}{\textbf{Delta Caption (\%)}} \\

& M & M & B & R & S & R & B & R & M \\
\midrule

No-Interaction (w/o PIT)
& 45.5 & 21.8 & 81.3 & 53.0 & 34.4 & 36.7 & 49.1 & 37.1 & 48.0 \\
 
Object-Level (Mean) 
& 52.9 & 32.3 & 81.0 & 49.7 & 25.0 & 23.7 & \textbf{51.7} & 34.7 & 49.6 \\

Object-Level (Max) 
& 56.6 & 36.5 & 81.3 & 45.1 & 23.9 & 22.5 & 50.9 & 37.2 & 50.0 \\
 
Micro-Token ($M=32$) 
& 56.9 & 35.3 & 80.1 & 44.0 & 24.5 & 21.8 & 50.6 & 34.6 & 50.0 \\
 
\midrule
\rowcolor[gray]{0.7}\textbf{w/ PIT} 
& 56.3 & 33.8 & \textbf{81.7} & \textbf{57.2} & \textbf{37.1} & \textbf{36.8} & 51.2 & \textbf{37.3} & \textbf{51.0} \\
\bottomrule
\end{tabular}
}
\end{table*}

\noindent\textbf{Results and analysis}
The results in Table~\ref{tab:ablation_interaction_extended} offer a nuanced and critical insight. On the semantic comparison tasks (MO3D), both \emph{Object-Level (Max)} and \emph{Micro-Token} variants perform exceptionally well, slightly surpassing our model with PIT block (Hereafter PIT model). This suggests that for high-level semantic comparison, capturing salient features via max-pooling or representative tokens is sufficient.
However, the results on Shape Mating reveal a fundamental limitation of these aggregation-based approaches. Both \emph{Object-Level (Max)} and \emph{Micro-Token} fail on this geometric task, scoring 23.9 and 24.5 on Selection (S), respectively. These scores are not only far below our \emph{PIT model} but also worse than the \emph{No-Interaction} baseline. 
This confirms that the failure of object-level models is not due to the pooling operation but stems from the architectural bottleneck of compressing local geometry into object-wise slots. Even with 32 micro-tokens, the spatial correspondence required for mating is lost.

\subsection{Training Strategy}
\label{ssec:ablation_training_stages}

\noindent\textbf{Motivation and Setup.}
Our training framework adopts a two-stage strategy. \emph{Phase 1 (Feature Alignment)} follows the PointLLM methodology~\cite{xu2024pointllm}, training only the projector on the 660K brief-description instructions from Objaverse-Cap3D to align point cloud features with the LLM's embedding space. This is followed by \emph{Phase 2 (Holistic Task-Mixture)}, which fine-tunes the Projector, PIT block, and LLM (while keeping the point encoder frozen) on our proposed benchmarks.

A natural question is whether Phase 1 is redundant: can the model learn to align modalities and reason about geometry simultaneously? To investigate this, we evaluate a \emph{1-Stage} variant. In this setting, we initialize the projector randomly and train the full model (Encoder, Projector, and LLM) directly on the holistic data mixture. We compare this against our standard \emph{2-Stage} approach.

\noindent\textbf{Results and Analysis.}
The results in Table~\ref{tab:ablation_training_stages} reveal a critical trade-off between task-specific optimization and general reasoning capability.
Interestingly, for our PIT model, the \emph{1-Stage} approach yields surprisingly high scores on the Mini-Applications. It achieves a Selection score of 66.6 on Shape Mating and a Delta Captioning score of 56.0. However, this comes at a severe cost: performance on the main MO3D benchmark drops significantly. Specifically, the Positional score falls from 56.3 to 47.6, and the Comparative score decreases from 33.8 to 24.5.
This suggests that without the initial alignment of Phase 1, the powerful PIT architecture tends to overfit to the specific templates and biases of the narrower Mini-App tasks, effectively becoming a task-specific specialist at the expense of general understanding. The \emph{1-Stage} model learns to exploit the limited linguistic patterns of Shape Mating but fails to ground the diverse, open-ended concepts required for MO3D.
In contrast, the \emph{2-Stage} approach ensures that the model is first grounded in a broad 3D-text semantic space. This pre-alignment acts as a necessary foundation, preventing the model from collapsing into task-specific shortcuts and enabling the robust, generalized comparison capabilities shown in the MO3D results. Thus, Phase 1 is essential for training a true generalist 3D-LLM.

\subsection{Robustness to Object Count (Scaling to 4--5 Objects)}
\label{ssec:scaling_objects}
While our standard dataset focuses on $n \in \{2, 3\}$ to maintain high token density for fine-grained geometric tasks, we investigated zero-shot extensions to scenes with 4 or 5 objects on the MO3D positional QA task. We evaluated two approaches:

\noindent\textbf{Naive Scaling.} When forcing $n=4$ or $5$ objects via Micro-Token compression to fit within the same fixed token budget, performance naturally drops on queries referencing the 4th or 5th objects (falling to 32\% and 14\%, respectively), indicating out-of-distribution difficulty and loss of fidelity.

\noindent\textbf{Test-Time Coarse-to-Fine.} Since many queries depend only on a small subset of objects, we apply a training-free, inference-time filtering strategy: (i) we extract referenced objects from the question via ordinal terms, (ii) add top-2 candidates using lightweight retrieval using CLIP~\cite{radford2021learning}, and (iii) remap the ordinals to this filtered subset before running Multi-3DLLM. This procedure successfully recovers reasoning capabilities, achieving 50\% and 51\% accuracy on 4-object and 5-object positional queries, respectively. This demonstrates a viable path for computation-efficient scaling despite LLM context limits.

\section{Advanced Analysis on Shape Mating}
\label{sec:advanced_shape_mating}

\subsection{Impact of Two-Turn Conversational Reasoning}
In the main paper (Table 1), we employed a strict single-turn generation protocol for the Shape Mating task. The model was required to output both the pair selection and a detailed geometric rationale in a single response (\eg, ``(1,3). Pair (1,3) can mate because...''). Under this constrained setting, Multi-3DLLM achieved a Selection accuracy of 37.1\%.

However, forcing a combined output creates a well-known objective imbalance during training. The Cross-Entropy (CE) loss becomes dominated by the long, generative rationale tokens, which inadvertently penalizes the short, categorical selection tokens. To mitigate this, we evaluated the model using a \emph{two-turn conversational (2-chat) approach}, inspired by the multi-turn capabilities of standard 3D-LLMs like PointLLM~\cite{xu2024pointllm}. The task is decoupled as follows:
\begin{itemize}
    \item \textbf{Turn 1 (Selection):} The user asks, ``Which pairs can mate? select one that applies.'' The model responds strictly with the selection, \eg, ``(1,3)''.
    \item \textbf{Turn 2 (Reasoning):} The user follows up with, ``Explain why.'' The model then generates the geometric rationale.
\end{itemize}
By decoupling the objective, the model can dedicate its full attention to the geometric matching in the first turn without the loss being diluted by the generation of long explanations. We utilize this optimized two-turn protocol to explore the model's real-world robustness in the following transfer experiments.

\subsection{Transfer to Real-World Scans}
To demonstrate that Shape Mating is not merely a synthetic procedural task, we evaluated the zero-shot transfer capability of our model on real-world scanned datasets: ScanObjectNN and OmniObject3D. 

A critical challenge in real-world transfer is the point density gap. Real-world scans often have severe point limitations (\eg, ScanObjectNN is limited to $\sim$2048 points), creating a density bottleneck when fed into an encoder pre-trained on 8,192 points. To match point densities, we augmented the evaluation with OmniObject3D.

Using the highly effective \emph{Two-Turn Conversational} prompt described above, we present the real-world transfer results in Table~\ref{tab:real_world_sm}. When point densities are properly matched, our model achieves a viable zero-shot selection accuracy of 36.3\% (well above the 25\% chance level). Furthermore, with a brief fine-tuning on just 5K real-world samples, the performance surges to 62.0\%. In stark contrast, the 2D-VLM baseline (LLaVA-7B) remains entirely at chance-level ($\sim$25\%) across all settings. This confirms that the Shape Mating task fundamentally preserves its geometry-centric nature across domain shifts, and our architecture maintains its robustness in real-world scenarios.

\begin{table}[t]
\centering
\caption{\textbf{Results on real-world scanned datasets.} Metric is \textbf{Selection Accuracy (S, \%)} following the main paper. Chance rate is 25\% for both.}
\label{tab:real_world_sm}
\footnotesize
\setlength{\tabcolsep}{2.5pt}
\renewcommand{\arraystretch}{0.85}
\begin{tabular}{lcc}
\toprule
Model & \makecell[c]{ScanObjectNN (Real)} & \makecell[c]{OmniObject3D (Real)} \\
\midrule
LLaVA-7B (Zero-shot) & 21.6 & 26.0 \\
\midrule
Ours (Zero-shot) & 29.0 & 36.3 \\
Ours (Fine-tune) & \textbf{48.0} & \textbf{62.0} \\
\bottomrule
\end{tabular}
\end{table}

\begin{table*}[t]
\centering
\caption{
    Ablation on training stages across architectures. We compare the \emph{1-Stage} and \emph{2-Stage} (Alignment $\rightarrow$ Holistic Tuning) strategies for both the \emph{No-Interaction} baseline and  full model \emph{w/ PIT}.
    The results investigate whether the initial feature alignment (Phase 1) is universally beneficial or specifically critical for our patch-interaction mechanism.
}
\label{tab:ablation_training_stages}
\resizebox{\textwidth}{!}{%
\begin{tabular}{l | c | c | cc | cc | cc | c}
\toprule
\multirow{4}{*}{\textbf{Model \& Training Strategy}} 
& \multicolumn{4}{c|}{\textbf{MO3D}} 
& \multicolumn{2}{c|}{\textbf{Shape Mating}} 
& \multicolumn{3}{c}{\textbf{Change Captioning}} \\ 
\cmidrule(lr){2-5} \cmidrule(lr){6-7} \cmidrule(lr){8-10} 
& \multicolumn{1}{c|}{\textbf{Positional (\%)}} 
& \multicolumn{1}{c|}{\textbf{Comparative (\%)}} 
& \multicolumn{2}{c|}{\textbf{Holistic (\%)}} 
& \multicolumn{2}{c|}{\textbf{Selection (\%)}} 
& \multicolumn{2}{c|}{\textbf{Verify (\%)}}   
& \multicolumn{1}{c}{\textbf{Delta Caption (\%)}} \\

& M & M & B & R & S & R & B & R & M \\
\midrule

No-Interaction (1-Stage)
& 47.6 & 21.0 & 78.9 & 41.9 & 23.8 & 22.0 & 49.1 & 39.4 & 42.0 \\
No-Interaction (2-Stage)
& 45.5 & 21.8 & 81.3 & 53.0 & 34.4 & 36.7 & 49.1 & 37.1 & 48.0 \\

\midrule
w/ PIT (1-Stage)
& 47.6 & 24.5 & 78.9 & 44.8 & \textbf{66.6} & \textbf{64.1} & 50.3 & 38.4 & \textbf{56.0} \\
\rowcolor[gray]{0.7}\textbf{w/ PIT (2-Stage)} 
& \textbf{56.3} & \textbf{33.8} & \textbf{81.7} & \textbf{57.2} & 37.1 & 36.8 & \textbf{51.2} & \textbf{37.3} & 51.0 \\

\bottomrule
\end{tabular}
}
\end{table*}

\section{Evaluation Details}
\label{sec:eval_details}

\subsection{LLM-based Evaluation Prompts}

To ensure a robust and semantic assessment, we utilize \textit{GPT-4o-mini} as our primary evaluator. Unlike rigid n-gram metrics, this LLM-based judge can discern semantic equivalence and validate reasoning logic. We employ specific prompts for each metric type, as detailed below and illustrated in Figure~\ref{fig:eval_prompts}.

\noindent\textbf{Semantic Accuracy (M) for MO3D.}
For open-ended questions in MO3D, exact string matching is insufficient. Our evaluation prompt instructs the judge to rate a response as Correct (1) or Incorrect (0). Crucially, this metric incorporates Visual Grounding. The evaluator is provided with both the ground-truth text and the multi-view images of the point clouds. It is instructed to accept the model's answer if it: (1) matches the ground truth semantically, or (2) provides a valid alternative description that is clearly supported by the visual evidence in the images, even if it differs from the text.

\noindent\textbf{Semantic Accuracy (M) for Delta Captioning.}
For the Change Captioning (Delta) task, a binary score is too coarse. We employ a 10-point scale prompt. The evaluator decomposes the ground-truth edit instruction into atomic components (\eg, ``thicker legs'', ``higher back'')  and grades the generated description based on the recall of these components. Contradictions (\eg, describing "thinner legs" when the truth is "thicker") result in an immediate score of 0.

\noindent\textbf{Reasoning Accuracy (R).}
For tasks requiring justification (Shape Mating, Change Captioning, and MO3D Holistic), we evaluate the quality of the ``Why'' output. The prompt provides the judge with the context (objects/instruction), the model's selected answer, and its reasoning. The judge assigns a score of 1 only if the reasoning is logically sound, factually consistent with the ground truth answer.

\noindent\textbf{Validation of the LLM Judge.}
To ensure the LLM metric is a reliable proxy for semantic evaluation, we conducted a human audit (3 people) of 300 randomly selected LLM-judged responses. The LLM's decisions achieved 91.0\% unanimous human support (and 98.0\% with at least one human vote). This confirms that our evaluation protocol accurately reflects human judgment in assessing 3D geometric descriptions and reasoning.

\subsection{Standard NLP Metrics}
\label{ssec:nlp_metrics}

We report standard NLP metrics \emph{BLEU-4}~\cite{papineni2002bleu}, \emph{ROUGE-L}~\cite{lin2004rouge, ganesan2018rouge}, \emph{METEOR}~\cite{banerjee2005meteor}, \emph{SimCSE}~\cite{gao2021simcse} for all generative tasks. 
Table~\ref{tab:nlp_metrics_miniapps} presents the results for the Mini-Applications (Change Captioning and Shape Mating), and Table~\ref{tab:nlp_metrics_mo3d} presents the results for the MO3D dataset. For 2D-VLMs, scores are explicitly reported for both 1-view and 2-view settings.

\noindent\textbf{Analysis of Metric Discrepancies.}
We observe instances where standard NLP metrics diverge from our semantic evaluators. For example, in Shape Mating, 2D-VLMs achieve high SimCSE scores (\eg, LLaVA: 67.65) despite near-zero Selection accuracy. This indicates ``hallucinated fluency'': generating plausible-sounding but geometrically incorrect text. Similarly, in MO3D Positional, the \emph{w/o PIT} baseline slightly edges out full Multi-3DLLM on BLEU-4, yet fails significantly on Semantic Accuracy (M). This suggests the baseline relies on memorizing safe linguistic patterns, whereas our model generates more diverse, geometrically grounded descriptions that differ from the ground truth text but are verified as correct by the LLM judge. These discrepancies underscore the necessity of our proposed LLM-based metrics for accurate benchmarking.

\begin{table*}[h]
\centering
\caption{
    \textbf{Standard NLP Metrics for MO3D.} 
    Detailed scores for Positional, Comparative, and Holistic QA tasks.
    SimCSE scores evaluate semantic similarity.
}
\label{tab:nlp_metrics_mo3d}
\resizebox{\textwidth}{!}{%
\begin{tabular}{l | cccc | cccc | cccc}
\toprule
\multirow{2}{*}{\textbf{Model}} 
& \multicolumn{4}{c|}{\textbf{Positional QA}} 
& \multicolumn{4}{c|}{\textbf{Comparative QA}} 
& \multicolumn{4}{c}{\textbf{Holistic QA}} \\
& \textbf{B-4}$\uparrow$ & \textbf{R-L}$\uparrow$ & \textbf{MET}$\uparrow$ & \textbf{Sim}$\uparrow$ 
& \textbf{B-4}$\uparrow$ & \textbf{R-L}$\uparrow$ & \textbf{MET}$\uparrow$ & \textbf{Sim}$\uparrow$ 
& \textbf{B-4}$\uparrow$ & \textbf{R-L}$\uparrow$ & \textbf{MET}$\uparrow$ & \textbf{Sim}$\uparrow$ \\
\midrule
LLaVA (1-view)
& 23.97 & 49.84 & 47.43 & 62.31 
& 3.97 & 26.48 & 28.66 & 58.46 
& 2.19 & 20.03 & 23.05 & 57.68 \\

LLaVA (2-view)
& 23.79 & 49.55 & 47.48 & 61.62 
& 3.59 & 26.18 & 27.52 & 58.16 
& 2.29 & 20.32 & 23.51 & 58.19 \\

Molmo (1-view)
& 3.88 & 24.27 & 31.26 & 55.04 
& 3.35 & 21.26 & 27.87 & 63.21 
& 1.01 & 15.15 & 19.92 & 54.22 \\

Molmo (2-view)
& 3.40 & 24.20 & 30.70 & 55.06 
& 3.40 & 21.23 & 27.43 & 62.79 
& 0.99 & 14.92 & 19.17 & 53.47 \\

\midrule
MiniGPT-3D
& 15.99 & 44.01 & 50.36 & 63.04 
& 13.73 & 42.01 & 40.52 & 64.59 
& 4.20 & 24.38 & 24.28 & 58.46 \\

PointLLM
& 29.95 & 55.60 & 52.22 & 64.65 
& 16.06 & 43.75 & 41.77 & 64.13 
& 3.60 & 21.32 & 21.97 & 61.26 \\

ShapeLLM
& 31.08 & 56.67 & 53.86 & 63.77 
& 24.23 & 49.68 & 48.56 & 71.43 
& 4.25 & 20.85 & 20.76 & 62.82 \\

\midrule
Multi-3DLLM (w/o PIT)
& \textbf{46.83} & \textbf{69.52} & 66.44 & \textbf{79.56} 
& 40.74 & 62.06 & 62.96 & 80.14 
& \textbf{26.87} & \textbf{52.34} & \textbf{49.84} & \textbf{76.22} \\

\rowcolor[gray]{0.7}\textbf{Multi-3DLLM (Ours)} 
& 45.54 & 68.95 & \textbf{66.59} & 78.32 
& \textbf{40.99} & \textbf{62.83} & \textbf{63.38} & 79.72 
& 22.79 & 49.25 & 47.40 & 74.52 \\

\bottomrule
\end{tabular}
}
\end{table*}

\begin{table*}[h]
\centering
\caption{
    \textbf{Standard NLP Metrics for Mini-Applications.} 
    Detailed scores for Shape Mating, Change Captioning (Verify), and Change Captioning (Delta). 
}
\label{tab:nlp_metrics_miniapps}
\resizebox{\textwidth}{!}{%
\begin{tabular}{l | cccc | cccc | cccc}
\toprule
\multirow{2}{*}{\textbf{Model}} 
& \multicolumn{4}{c|}{\textbf{Shape Mating}} 
& \multicolumn{4}{c|}{\textbf{Change Captioning (Verify)}} 
& \multicolumn{4}{c}{\textbf{Change Captioning (Delta)}} \\
& \textbf{B-4}$\uparrow$ & \textbf{R-L}$\uparrow$ & \textbf{MET}$\uparrow$ & \textbf{Sim}$\uparrow$ 
& \textbf{B-4}$\uparrow$ & \textbf{R-L}$\uparrow$ & \textbf{MET}$\uparrow$ & \textbf{Sim}$\uparrow$ 
& \textbf{B-4}$\uparrow$ & \textbf{R-L}$\uparrow$ & \textbf{MET}$\uparrow$ & \textbf{Sim}$\uparrow$ \\
\midrule
LLaVA (1-view)
& 1.17 & 21.99 & 15.31 & 67.48 
& 4.53 & 30.05 & 23.57 & 56.27 
& 3.58 & 26.69 & 19.85 & 55.86 \\

LLaVA (2-view)
& 1.17 & 22.23 & 15.42 & \textbf{67.81}
& 4.52 & 28.51 & 22.76 & 55.39 
& 3.33 & 26.26 & 18.77 & 56.28 \\

Molmo (1-view)
& 2.59 & 24.66 & 20.60 & 57.78 
& 1.98 & 21.11 & 24.29 & 55.84 
& 1.52 & 18.26 & 21.49 & 52.71 \\

Molmo (2-view)
& 2.85 & 23.26 & 20.60 & 60.45 
& 2.02 & 21.50 & 24.77 & 55.76 
& 1.49 & 18.24 & 21.73 & 52.78 \\

\midrule
MiniGPT-3D
& 1.91 & 14.93 & 15.78 & 45.86 
& 2.60 & 22.12 & 15.87 & 45.47 
& 0.98 & 11.95 & 7.76 & 33.75 \\

PointLLM
& 1.13 & 13.04 & 12.61 & 42.74 
& 1.76 & 14.22 & 13.62 & 38.29 
& 0.72 & 11.90 & 8.88 & 39.90 \\

ShapeLLM
& 1.73 & 16.58 & 13.47 & 43.70 
& 1.33 & 9.60 & 10.44 & 23.68 
& 1.23 & 15.09 & 13.92 & 48.21 \\

\midrule
Multi-3DLLM (w/o PIT)
& 16.28 & 31.43 & 28.19 & 53.28 
& 10.40 & 34.67 & 29.20 & 55.66 
& 5.00 & 25.98 & \textbf{24.14} & \textbf{68.42} \\

\rowcolor[gray]{0.7}\textbf{Multi-3DLLM (Ours)} 
& \textbf{16.65} & \textbf{31.40} & \textbf{28.37} & 53.23 
& \textbf{12.47} & \textbf{37.67} & \textbf{30.76} & \textbf{57.62} 
& \textbf{5.02} & \textbf{26.59} & 23.98 & 67.60 \\

\bottomrule
\end{tabular}
}
\end{table*}

\section{Additional Qualitative Results}
\label{sec:qualitative}

We provide extensive qualitative examples to visually demonstrate the capabilities and limitations of our model compared to state-of-the-art baselines.

\noindent\textbf{Comparison on MO3D (Main Task).}
Figures~\ref{fig:qual_mo3d_pos_1} to \ref{fig:qual_mo3d_hol_2} (Examples 1--5) present results comparisons on the MO3D benchmark.

\noindent\textbf{Performance on Mini-Applications.}
Figures~\ref{fig:qual_sm_1} to \ref{fig:qual_cc_delta} (Examples 6--9) showcase results on the application-driven benchmarks.

\begin{figure*}[h]
    \centering
    \begin{tcolorbox}[
        colback=white,
        colframe=gray!50!black,
        title=\textbf{\sffamily System Prompt for MO3D QA Generation},
        fonttitle=\bfseries\large,
        boxrule=1pt,
        arc=2mm,
        left=10pt, right=10pt, top=10pt, bottom=10pt
    ]
    \small
    \ttfamily
    
    You are an AI assistant creating a high-quality dataset for a 3D vision-language model. 
    Your task is to generate question–answer pairs for three task types (positional, comparative, holistic) 
    based on the object descriptions below. (and supplementary multi-view images when available). Use visual evidence as the primary source for visual attributes. Use descriptions to supplement non-visual semantics. When images contradict the descriptions on visual attributes, trust the images. Do not claim uniqueness from omission in descriptions; verify across images.

    \vspace{0.6em}
    Input Context: \textit{[Object Descriptions \& Multi-view Images]}

    \vspace{0.6em}
    CRITICAL INSTRUCTIONS:
    \begin{enumerate}[leftmargin=2em, label=\arabic*., nosep]
        \item Produce exactly two distinct question–answer pairs for each task type.
        \item Answers must be direct, factual statements grounded in the images.
        \item Do NOT use object category nouns; instead refer to “the first object”, etc.
        \item Do NOT use spatial relations (left/right/front/behind).
        \item Comparative QA must emphasize structural or functional differences.
        \item Holistic QA must include one “Yes” and one “No” answer.

        \item MANDATORY Category-Specific Questions: 
        All questions must explicitly reference the target category (e.g., \texttt{\{target\_category\}}).  
        The question text must use the category’s REQUIRED KEYWORDS to avoid ambiguity.
        
        \vspace{0.4em}
        REQUIRED KEYWORDS (Excerpt):
        \begin{itemize}[leftmargin=1.5em, label=--]
            \item Geometry / Structure:  
            \hspace*{1em}MUST use: “shape”, “form”, “geometric”, “structural design”  
            \hspace*{1em}FORBIDDEN: “feature”, “property”, “appearance”
            
            \item Material:
            \hspace*{1em}MUST use: “material”, “made of”, “constructed from”  
            \hspace*{1em}FORBIDDEN: “feature”, “property”, “what is X”
            
            \item \textit{[Additional mandatory keyword rules for: Color, Function, Taxonomy, Style/Aesthetics]}
        \end{itemize}
        \vspace{0.4em}

        \item Within each task, the two variants must rely on different properties.
        \item Do NOT assume a property is unique unless the images clearly show uniqueness.
        \item For positional questions referencing a “unique” feature, ensure exactly one object has that feature.
    \end{enumerate}

    \vspace{0.6em}
    Task Definitions:
    \begin{itemize}[leftmargin=2em, nosep, label=--]
        \item positional\_qa: Question about one object’s attribute.
        \item comparative\_qa: Question comparing structural or functional aspects.
        \item holistic\_qa: Question about a property shared (or not shared) by all objects.
    \end{itemize}

    \end{tcolorbox}

    \caption{\textbf{System Prompt for MO3D QA Generation.} 
To ensure high-quality, non-ambiguous questions, we enforce strict keyword constraints (Instruction 7) for each target category. 
For readability, we list representative keywords for the "Geometry" and "Material" categories; identical constraint logic is applied to the other categories (Color, Function, Taxonomy, Style).}
    \label{fig:prompt_mo3d}
\end{figure*}

\begin{figure*}[t]
    \centering
    \begin{tcolorbox}[colback=gray!5!white,colframe=gray!75!black,title=\textbf{System Prompt for Shape Mating Rationale Paraphrasing}]
    \small
    \fontfamily{pcr}\selectfont
    You are rewriting rationales for a 3D part mating QA dataset.
    Three interface parts are labeled (1), (2), and (3).
    The correct mating pair list must remain \texttt{[Answer List]}.
    The canonical rationales are:
    \texttt{[Canonical Rationales]}
    
    Task: Paraphrase each rationale in fresh wording while keeping the facts.
    
    Guidelines:
    \begin{enumerate}[leftmargin=*, nosep]
        \item Preserve the logical meaning of every rationale.
        \item Mention complementary vs. conflicting geometry explicitly.
        \item Keep each entry to one or two sentences. Vary phrasing in a \texttt{[Style Hint]} style.
        \item Do not introduce new geometry details or contradict the canonical text.
        \item Return JSON only, with keys 'answer' and 'why'.
        \begin{itemize}[nosep]
            \item 'answer' must be the same list of mating pairs.
            \item 'why' must map each pair key to your rewritten rationale.
        \end{itemize}
        \item Use the exact keys '(1,2)', '(1,3)', '(2,3)'.
        \item No code fences, no additional commentary.
    \end{enumerate}
    \end{tcolorbox}
    \caption{\textbf{Full Prompt for Shape Mating Rationale Generation.} We utilize GPT-4o-mini to convert structured error tags (e.g., \texttt{cut\_mismatch}) into natural language explanations.}
    \label{fig:prompt_sm}
\end{figure*}

\begin{figure*}[t]
    \centering
    \begin{tcolorbox}[colback=gray!5!white,colframe=gray!75!black,title=\textbf{Templates for Change Captioning Tasks}]
    \small
    \fontfamily{pcr}\selectfont
    
    \textbf{Task 1: Verification (Binary Classification)}
    \vspace{2mm}
    
    \textit{Input:} Anchor Point Cloud ($P_A$), Candidate Point Cloud ($P_C$), Instruction ($I$)
    \newline
    \textit{Randomization:} The order of input point clouds ($P_A, P_C$) is randomized.
    \vspace{1mm}
    
    [Template A: Anchor is First]
    \newline
    Input: \texttt{<point> ($P_A$)} \texttt{<point> ($P_C$)}
    \newline
    Q: \texttt{Does the second object satisfy all of the following requirements compared to the first object?}
    \newline
    \texttt{Requirements: - [Instruction]}
    \vspace{1mm}
    
    [Template B: Anchor is Second]
    \newline
    Input: \texttt{<point> ($P_C$)} \texttt{<point> ($P_A$)}
    \newline
    Q: \texttt{Does the first object satisfy all of the following requirements compared to the second object?}
    \newline
    \texttt{Requirements: - [Instruction]}
    
    \vspace{3mm}
    \hrule
    \vspace{3mm}
    
    \textbf{Task 2: Delta Captioning (Generative)}
    \vspace{2mm}
    
    \textit{Input:} Anchor Point Cloud ($P_A$), Positive Point Cloud ($P_P$)
    \newline
    \textit{Goal:} Generate a description of the geometric edit.
    \vspace{1mm}
    
    [Templates (Randomly Selected)]
    \begin{itemize}[leftmargin=*, nosep]
        \item \texttt{How would you transform the first object so that it matches the second object?}
        \item \texttt{Describe the edits needed to convert the first object into the second.}
        \item \texttt{What modifications should be applied to the first object to obtain the second?}
        \item \texttt{List the geometric adjustments required to turn the first object into the second.}
    \end{itemize}
    \end{tcolorbox}
    \caption{\textbf{Prompt Templates for Change Captioning.} We utilize a set of diverse templates for the Delta Captioning task. For the Verification task, we explicitly randomize the input order of the anchor and candidate objects and adjust the question wording ("first" vs "second" object) accordingly to prevent positional bias.}
    \label{fig:prompt_cc}
\end{figure*}

\begin{table*}[h]
\centering
\small
\setlength{\tabcolsep}{10pt}
\renewcommand{\arraystretch}{1.2}
\caption{
    \textbf{Instruction Templates for Shape Mating.} 
    We utilize 15 variations of the prompt to train the model, ensuring robustness to phrasing while maintaining a consistent task definition (Selection + Explanation).
}
\label{tab:sm_templates}
\begin{tabular}{p{0.95\linewidth}}
\toprule
\textbf{Template Variations} \\
\midrule
\textbullet\ Which pairs can mate? select one that applies.\newline Options: (1,2), (1,3), (2,3), None\newline Explain why each chosen pair can mate and why others cannot. \\
\midrule
\textbullet\ Identify which pairs can mate. Choose the applicable option.\newline Options: (1,2), (1,3), (2,3), None\newline Provide explanations for why each selected pair can mate and why the others cannot. \\
\midrule
\textbullet\ Determine which pairs are able to mate. select one that applies.\newline Options: (1,2), (1,3), (2,3), None\newline Explain the reasoning for each pair that can mate and why the remaining pairs cannot. \\
\midrule
\textbullet\ Find the pairs that can mate together. Select the applicable option.\newline Options: (1,2), (1,3), (2,3), None\newline Describe why each chosen pair can mate and explain why the other pairs cannot. \\
\midrule
\textbullet\ Which pairs can successfully mate? select one that applies.\newline Options: (1,2), (1,3), (2,3), None\newline Explain why each selected pair can mate and provide reasons why the others cannot. \\
\midrule
\textbullet\ Determine all pairs that are capable of mating. Select the applicable options.\newline Options: (1,2), (1,3), (2,3), None\newline Provide explanations for why each selected pair can mate and why others cannot. \\
\midrule
\textbullet\ Which pairs can mate with each other? select one that applies.\newline Options: (1,2), (1,3), (2,3), None\newline Explain why each chosen pair can mate and describe why the other pairs cannot. \\
\midrule
\textbullet\ What pairs are able to mate? select one that applies.\newline Options: (1,2), (1,3), (2,3), None\newline Explain why each selected pair can mate and justify why the other pairs cannot. \\
\midrule
\textbullet\ Find all pairs that can mate. Choose  applicable option.\newline Options: (1,2), (1,3), (2,3), None\newline Describe why each chosen pair can mate and explain why the remaining pairs cannot. \\
\midrule
\textbullet\ Which pairs can be mated together? select one that applies.\newline Options: (1,2), (1,3), (2,3), None\newline Explain why each selected pair can mate and provide reasoning for why others cannot. \\
\midrule
\textbullet\ Identify the pairs that can mate. Choose the applicable option.\newline Options: (1,2), (1,3), (2,3), None\newline Explain why each chosen pair can mate and describe why the other pairs cannot. \\
\midrule
\textbullet\ Determine which pairs are compatible for mating. select one that applies.\newline Options: (1,2), (1,3), (2,3), None\newline Provide explanations for why each selected pair can mate and why the remaining pairs cannot. \\
\midrule
\textbullet\ What pairs can successfully mate? select one that applies.\newline Options: (1,2), (1,3), (2,3), None\newline Explain why each chosen pair can mate and justify why others cannot. \\
\midrule
\textbullet\ Which pairs are capable of mating? select one that applies.\newline Options: (1,2), (1,3), (2,3), None\newline Explain why each selected pair can mate and provide reasons why the other pairs cannot. \\
\bottomrule
\end{tabular}
\end{table*}

\begin{table*}[h]
\centering
\small
\caption{
    \textbf{Instruction Templates for Change Captioning.} 
    We use distinct template sets for the Verification and Delta Captioning tasks. 
    For Verification, we explicitly randomize the object order (Anchor first vs. second) to prevent positional bias.
}
\label{tab:cc_templates}
\begin{tabular}{l p{0.85\linewidth}}
\toprule
\textbf{Task Type} & \textbf{Template Variations} \\
\midrule
\textbf{Verification} 
& \textbullet\ \textit{(Condition: Anchor is First object):} \newline Does the second object satisfy all of the following requirements compared to the first object?\newline Requirements: - [Instruction] \\
& \textbullet\ \textit{(Condition: Anchor is Second object):} \newline Does the first object satisfy all of the following requirements compared to the second object?\newline Requirements: - [Instruction] \\
\midrule
\textbf{Delta Captioning} 
& \textbullet\ How would you transform the first object so that it matches the second object? \\
& \textbullet\ Describe the edits needed to convert the first object into the second. \\
& \textbullet\ What modifications should be applied to the first object to obtain the second? \\
& \textbullet\ List the geometric adjustments required to turn the first object into the second. \\
\bottomrule
\end{tabular}
\end{table*}

\begin{figure*}[t]
    \centering
\includegraphics[width=1.0\linewidth]{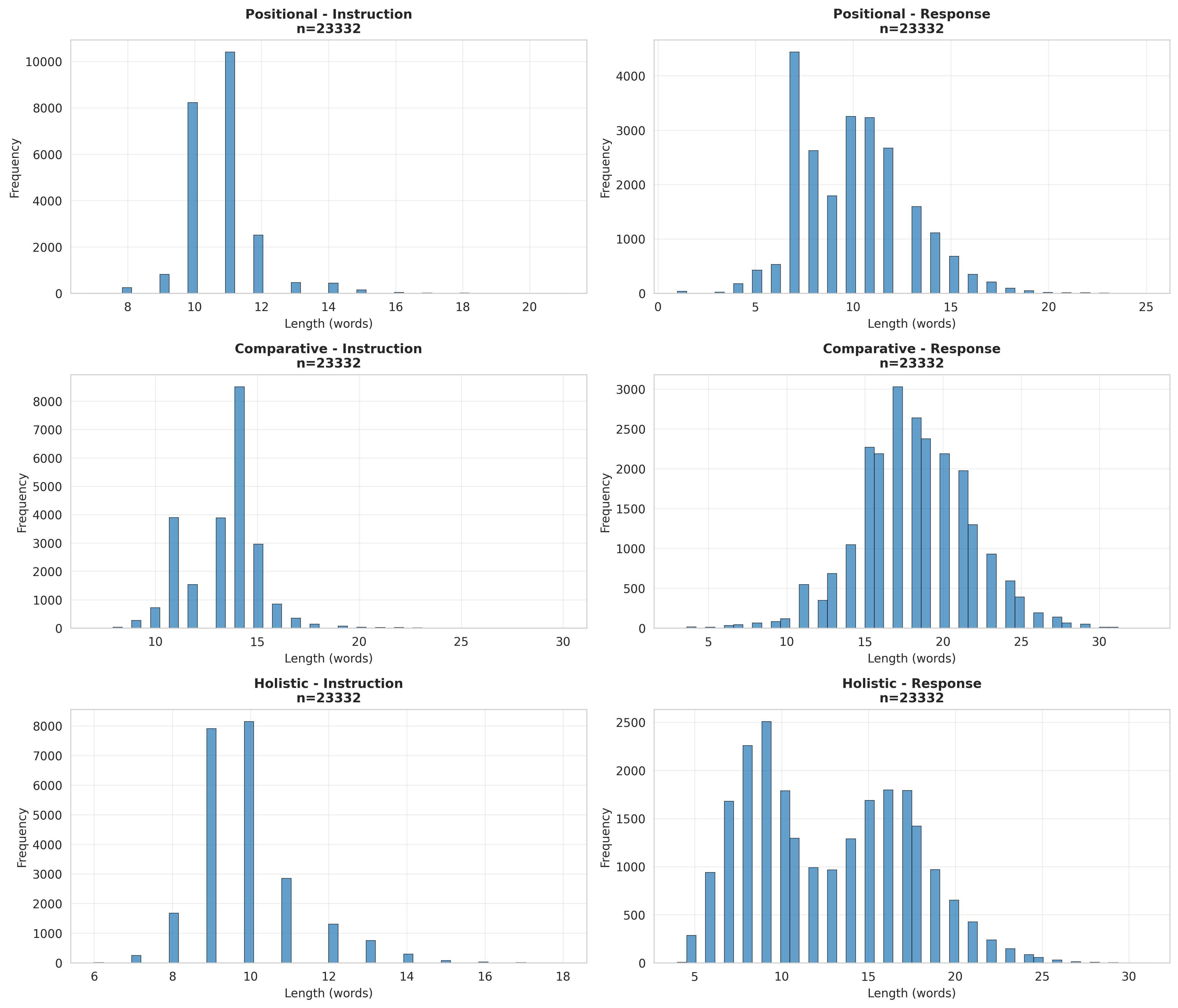}
    \caption{
        \textbf{Distribution of instruction and response lengths in MO3D dataset.}
    }
    \label{fig:mo3d_length_dist}
\end{figure*}

\begin{figure*}[t]
    \centering
\includegraphics[width=1.0\linewidth]{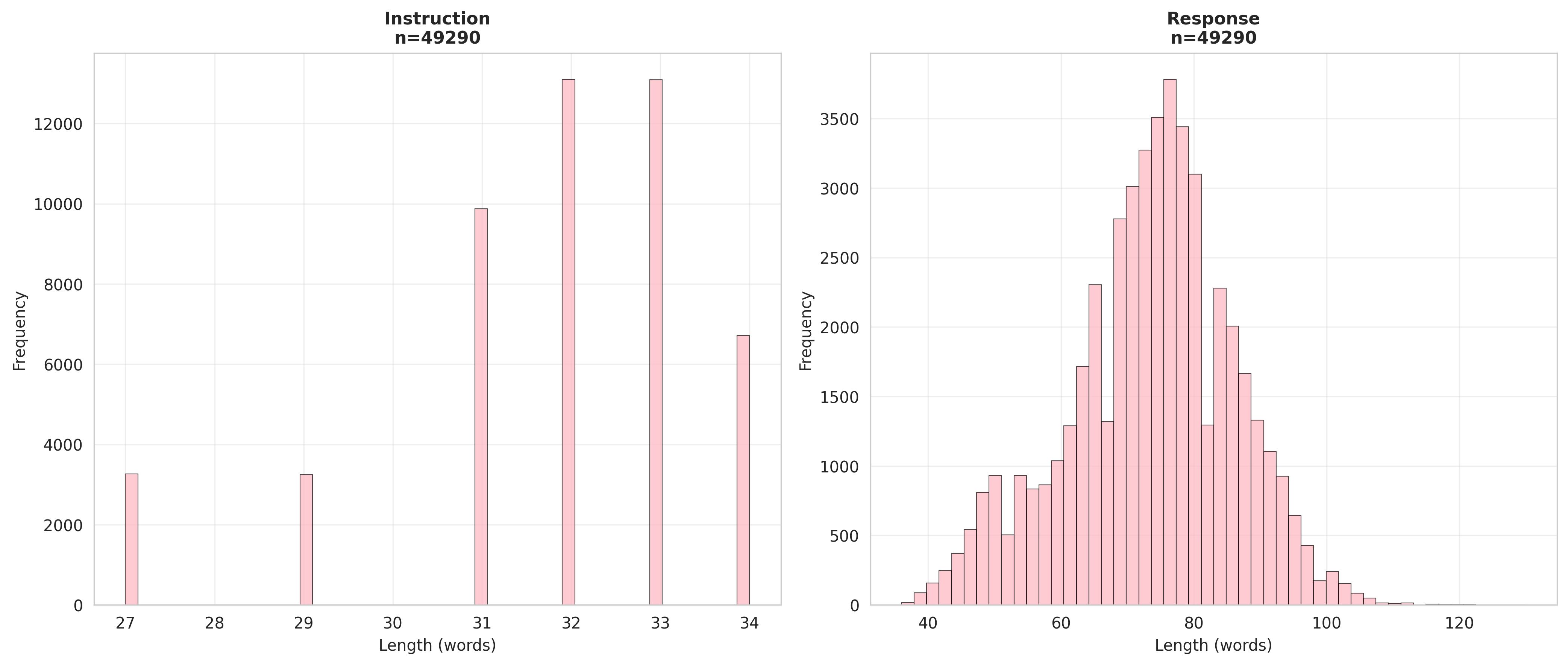}
    \caption{
        \textbf{Distribution of instruction and response lengths in Shape Mating.}
    }
    \label{fig:sm_length_dist}
\end{figure*}

\begin{figure*}[t]
    \centering
\includegraphics[width=1.0\linewidth]{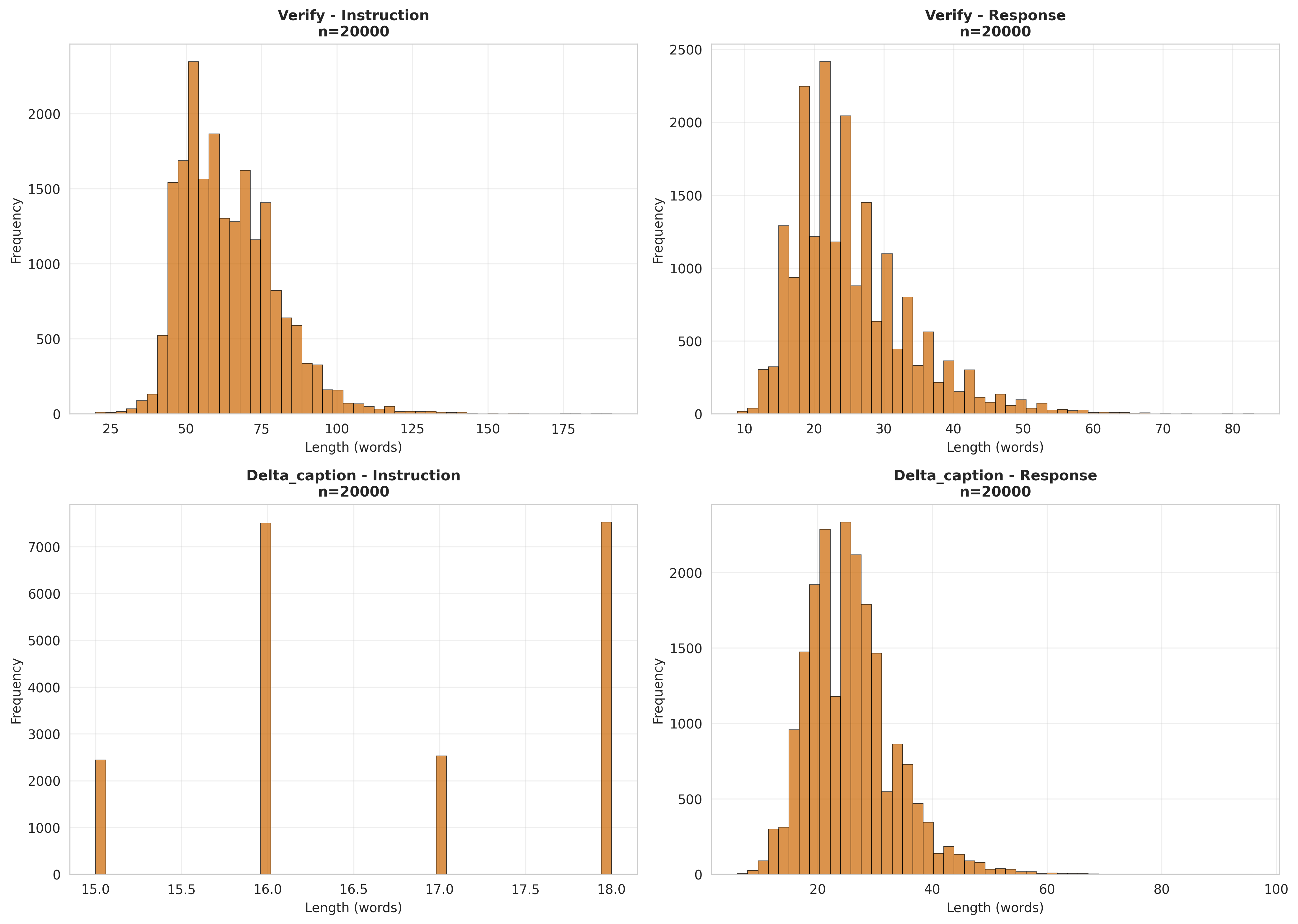}
    \caption{
        \textbf{Distribution of instruction and response lengths in Change Captioning.}
    }
    \label{fig:cc_length_dist}
\end{figure*}

\begin{figure*}[t]
    \centering
    \begin{tcolorbox}[colback=gray!5!white,colframe=gray!75!black,title=\textbf{LLM Evaluation Prompts (GPT-4o-mini)}]
    \small
    \fontfamily{pcr}\selectfont
    
    \textbf{[Metric M: Semantic Accuracy for MO3D]}
    \newline
    System: You are an impartial grader for a 3D-QA benchmark with multi-view image evidence.
    \newline
    CRITICAL: The images are the PRIMARY source of truth. Ground-truth text is a reference.
    \newline
    Criteria:
    \begin{enumerate}[leftmargin=*, nosep]
        \item Accept the answer if it matches the ground-truth text semantically.
        \item Accept the answer if it reasonably describes what is visible in the images, even if it differs from the ground-truth text (e.g., specific material/shape details).
        \item Only reject if the answer clearly contradicts what is visible in ALL provided images.
    \end{enumerate}
    Output JSON: \texttt{\{"score": 1 or 0, "reason": "..."\}}
    
    \vspace{2mm}
    \hrule
    \vspace{2mm}
    
    \textbf{[Metric M: Delta Captioning Score (10-point scale)]}
    \newline
    System: Evaluate the model's description of geometric changes using a 10-point scale.
    \newline
    Instructions:
    \begin{enumerate}[leftmargin=*, nosep]
        \item Break down the ground truth into individual geometric modification items.
        \item Check how many items are captured by the model.
        \item If the model contradicts any item, return M=0.
        \item Score based on coverage: 10 (All items correct), 7-9 (Most correct), 4-6 (Half correct), 1-3 (Few correct), 0 (Contradiction/None).
    \end{enumerate}
    \textbf{Output JSON:} \texttt{\{"M": 0-10, "reason": "..."\}}
    
    \vspace{2mm}
    \hrule
    \vspace{2mm}
    
    \textbf{[Metric R: Reasoning Accuracy]}
    \newline
    System: Evaluate whether the model's reasoning is factually consistent with the requirements and ground truth.
    \newline
    Criteria:
    \begin{itemize}[leftmargin=*, nosep]
        \item Is the reasoning logically sound?
        \item Does it correctly justify the selected answer/conclusion?
    \end{itemize}
    Output JSON: \texttt{\{"R": 1 or 0, "reason": "..."\}}
    \end{tcolorbox}
    \caption{\textbf{System Prompts for Evaluation.} We use specific prompts for different metric types. For MO3D (Top), the evaluator is explicitly instructed to prioritize visual evidence from multi-view renderings over text labels to handle valid open-ended descriptions.}
    \label{fig:eval_prompts}
\end{figure*}

\begin{figure*}[p]
    \centering
    \small 
    \setlength{\tabcolsep}{6pt}
    \renewcommand{\arraystretch}{1.3}

    \noindent\textbf{Example 1: Positional Understanding}
    \vspace{2mm}

    \includegraphics[width=0.95\linewidth, height=7cm, keepaspectratio]{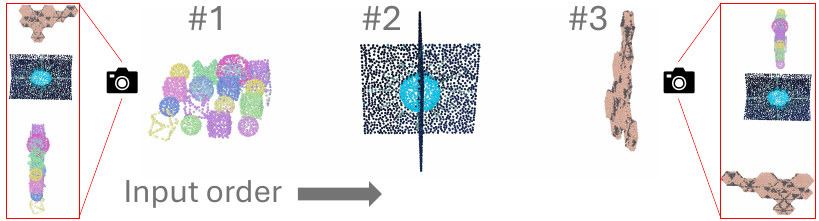} 
    
    \vspace{3mm}
    
    \begin{minipage}{0.95\textwidth}
        \textbf{Question:} What is the geometric shape of the third object? \\
        \textbf{Ground Truth:} The third object has a hexagonal geometric shape.
        \vspace{2mm}
        
        \begin{tabular}{l | p{0.85\textwidth}}
            \toprule
            \textbf{Model} & \textbf{Response} \\
            \midrule
            LLaVA (1-view)~\cite{liu2023visual} & {The third object is a square.} \\
            \midrule
            Molmo (1-view)~\cite{deitke2025molom} & {The geometric shape of the third object is a cross. This cross shape is formed by four rectangular prisms extending outward from a central point, creating a symmetrical cross-like structure.} \\
            \midrule
            MiniGPT-3D (Concat)~\cite{tang2024minigpt3d} & {the third object is a cube.} \\
            \midrule
            PointLLM (Concat)~\cite{xu2024pointllm} & {The third object is in the shape of a cube.} \\
            \midrule
            ShapeLLM (Concat)~\cite{qi2024shapellm} & {The third object is a cube.} \\
            \midrule
            \rowcolor{green!10}
            \textbf{Multi-3DLLM (Ours)} & \textbf{The third object has a hexagonal geometric shape.} \\
            \bottomrule
        \end{tabular}
    \end{minipage}

    \vspace{5mm}

    \caption{
        \textbf{Qualitative Comparison on MO3D (Positional Task).} 
    }
    \label{fig:qual_mo3d_pos_1}
\end{figure*}

\begin{figure*}[p]
    \centering
    \small 
    \setlength{\tabcolsep}{6pt}
    \renewcommand{\arraystretch}{1.3}

    \noindent\textbf{Example 2: Positional Understanding}
    \vspace{2mm}

    \includegraphics[width=0.95\linewidth, height=7cm, keepaspectratio]{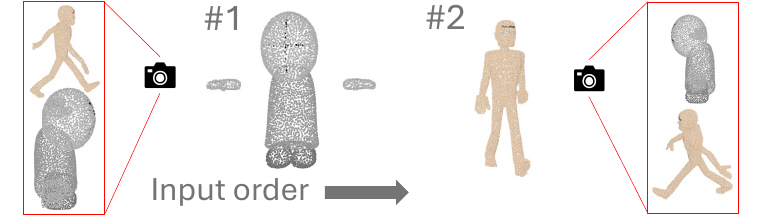} 
    
    \vspace{3mm}
    
    \begin{minipage}{0.95\textwidth}
        \textbf{Question:} What is the structural design of the second object? \\
        \textbf{Ground Truth:} The second object has a humanoid form with limbs positioned in a walking pose.
        \vspace{2mm}
        
        \begin{tabular}{l | p{0.85\textwidth}}
            \toprule
            \textbf{Model} & \textbf{Response} \\
            \midrule
            LLaVA (1-view)~\cite{liu2023visual} & {The second object is a cartoon-like figure, which is a representation of a person walking.} \\
            \midrule
            Molmo (1-view)~\cite{deitke2025molom} & {The second object in the image is a human figure with a unique structural design. It has a simple, stylized form with a round head and a single, continuous line that forms the body, arms, and legs. The figure is depicted in a walking pose, with the left leg extended forward and the right leg behind. The arms are positioned in a way that suggests movement, with the left arm bent at the elbow and the right arm extended backward. The figure's skin tone is a light brown color, and there's a small black area above the left eye, which could represent an eye or some other facial feature. This design creates a minimalist yet dynamic representation of a human in motion.} \\
            \midrule
            MiniGPT-3D (Concat)~\cite{tang2024minigpt3d} & {the second object is a humanoid figure with a distinct, elongated torso and a head that is significantly larger than the rest of its body.} \\
            \midrule
            PointLLM (Concat)~\cite{xu2024pointllm} & {The second object is designed to resemble an arrow.} \\
            \midrule
            ShapeLLM (Concat)~\cite{qi2024shapellm} & {The second object has a cylindrical body with a flat base and two protruding arms or legs.} \\
            \midrule
            \rowcolor{green!10}
            \textbf{Multi-3DLLM (Ours)} & \textbf{The second object has a humanoid form with outstretched arms.} \\
            \bottomrule
        \end{tabular}
    \end{minipage}

    \vspace{5mm}

    \caption{
        \textbf{Qualitative Comparison on MO3D (Positional Task).} 
    }
    \label{fig:qual_mo3d_pos_2}
\end{figure*}

\begin{figure*}[p]
    \centering
    \small 
    \setlength{\tabcolsep}{6pt}
    \renewcommand{\arraystretch}{1.3}

    \noindent\textbf{Example 3: Comparison Understanding}
    \vspace{2mm}

    \includegraphics[width=0.95\linewidth, height=7cm, keepaspectratio]{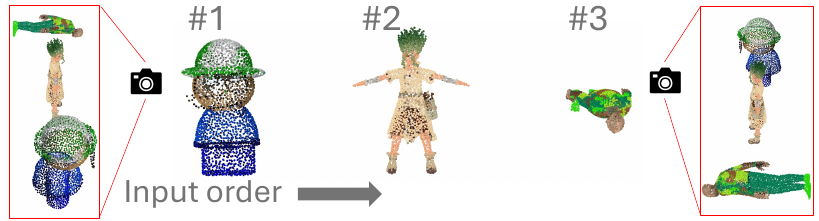} 
    
    \vspace{3mm}
    
    \begin{minipage}{0.95\textwidth}
        \textbf{Question:} How do the colors of the first and third objects compare? \\
        \textbf{Ground Truth:} The first object is primarily blue and green, while the third object features a camouflage pattern with green and brown shades.
        \vspace{2mm}
        
        \begin{tabular}{l | p{0.85\textwidth}}
            \toprule
            \textbf{Model} & \textbf{Response} \\
            \midrule
            LLaVA (1-view)~\cite{liu2023visual} & {The first object, which is a green hat, has a similar color to the third object, which is a green shirt. The hat and shirt both have a green hue, making them visually similar. The second object, a blue hat, has a distinct color compared to the green hat and shirt, as it is a different shade of blue.} \\
            \midrule
            Molmo (1-view)~\cite{deitke2025molom} & {The first and third objects in the image share the same color scheme. Both are wearing green pants. The first object has a green and yellow shirt, while the third object has a green shirt with yellow and red accents. So while the shirts differ in their patterns and additional colors, the base color for both the pants and shirts in the first and third objects is green.} \\
            \midrule
            MiniGPT-3D (Concat)~\cite{tang2024minigpt3d} & {the first object, a green hat, contrasts with the blue shirt of the second object.} \\
            \midrule
            PointLLM (Concat)~\cite{xu2024pointllm} & {The first and third objects have similar colors - both are green and blue.} \\
            \midrule
            ShapeLLM (Concat)~\cite{qi2024shapellm} & {The first object is green, while the third object is blue.} \\
            \midrule
            \rowcolor{green!10}
            \textbf{Multi-3DLLM (Ours)} & \textbf{The first object has a green hat and blue glasses, while the third object is primarily green with yellow accents.} \\
            \bottomrule
        \end{tabular}
    \end{minipage}

    \vspace{5mm}

    \caption{
        \textbf{Qualitative Comparison on MO3D (Comparison Task).} 
    }
    \label{fig:qual_mo3d_comp_1}
\end{figure*}

\begin{figure*}[p]
    \centering
    \small 
    \setlength{\tabcolsep}{6pt}
    \renewcommand{\arraystretch}{1.3}

    \noindent\textbf{Example 4: Holistic Understanding}
    \vspace{2mm}

    \includegraphics[width=0.95\linewidth, height=7cm, keepaspectratio]{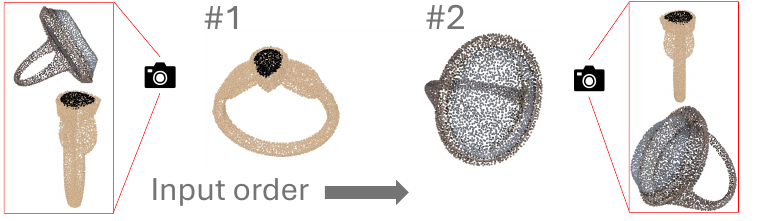} 
    
    \vspace{3mm}
    
    \begin{minipage}{0.95\textwidth}
        \textbf{Question:} Do both objects have a central gem or lens? \\
        \textbf{Ground Truth:} Yes. Both have a central element, one being a black diamond and the other a clear glass lens.
        \vspace{2mm}
        
        \begin{tabular}{l | p{0.85\textwidth}}
            \toprule
            \textbf{Model} & \textbf{Response} \\
            \midrule
            LLaVA (1-view)~\cite{liu2023visual} & {Yes, both objects have a central gem or lens.} \\
            \midrule
            Molmo (1-view)~\cite{deitke2025molom} & {No. Only the left object has a central gem or lens. The right object appears to be a simple ring without any central feature.} \\
            \midrule
            MiniGPT-3D (Concat)~\cite{tang2024minigpt3d} & {no, only the ring has a central gem.} \\
            \midrule
            PointLLM (Concat)~\cite{xu2024pointllm} & {Yes, both objects have a central gem or lens} \\
            \midrule
            ShapeLLM (Concat)~\cite{qi2024shapellm} & {Yes, both objects have a central gem or lens.} \\
            \midrule
            \rowcolor{green!10}
            \textbf{Multi-3DLLM (Ours)} & \textbf{Yes. Both objects feature a central diamond or lens.} \\
            \bottomrule
        \end{tabular}
    \end{minipage}

    \vspace{5mm}

    \caption{
        \textbf{Qualitative Comparison on MO3D (Holistic Task).} 
    }
    \label{fig:qual_mo3d_hol_1}
\end{figure*}

\begin{figure*}[p]
    \centering
    \small 
    \setlength{\tabcolsep}{6pt}
    \renewcommand{\arraystretch}{1.3}

    \noindent\textbf{Example 5: Holistic Understanding}
    \vspace{2mm}

    \includegraphics[width=0.95\linewidth, height=7cm, keepaspectratio]{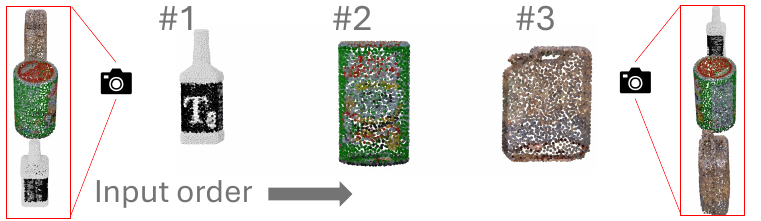} 
    
    \vspace{3mm}
    
    \begin{minipage}{0.95\textwidth}
        \textbf{Question:} Do all objects have a cylindrical form? \\
        \textbf{Ground Truth:} No. Only the second object has a cylindrical form, while the first object is a bottle and the third object is a canister.
        \vspace{2mm}
        
        \begin{tabular}{l | p{0.85\textwidth}}
            \toprule
            \textbf{Model} & \textbf{Response} \\
            \midrule
            LLaVA (1-view)~\cite{liu2023visual} & {No, not all objects have a cylindrical form. In the image, there is a bottle, a can, and a vase, which are all cylindrical in shape. However, there is also a box, which is not cylindrical.} \\
            \midrule
            Molmo (1-view)~\cite{deitke2025molom} & {No, not all objects in the image have a cylindrical form. While there is a prominent cylindrical structure in the center, which appears to be a stone or concrete pillar with a square top, there are other objects visible that have different shapes. In the background, you can see a rectangular object, and there's also a triangular shape visible. These various shapes create an interesting contrast with the main cylindrical structure, adding depth and diversity to the overall composition of the image.} \\
            \midrule
            MiniGPT-3D (Concat)~\cite{tang2024minigpt3d} & {no, not all objects have a cylindrical form. the objects in the 3d model are a bottle, and a can.} \\
            \midrule
            PointLLM (Concat)~\cite{xu2024pointllm} & {No, while two objects do have a cylindrical form, the third one is in the shape of a cuboid.} \\
            \midrule
            ShapeLLM (Concat)~\cite{qi2024shapellm} & {No, not all objects have a cylindrical form.} \\
            \midrule
            \rowcolor{green!10}
            \textbf{Multi-3DLLM (Ours)} & \textbf{Yes. All objects exhibit a cylindrical shape.} \\
            \bottomrule
        \end{tabular}
    \end{minipage}

    \vspace{5mm}

    \caption{
        \textbf{Qualitative Comparison on MO3D (Holistic Task).} (Note: This is a failure case where our model incorrectly generalizes the cylindrical shape to all objects.) 
    }
    \label{fig:qual_mo3d_hol_2}
\end{figure*}

\begin{figure*}[p]
    \centering
    \small 
    \setlength{\tabcolsep}{6pt}
    \renewcommand{\arraystretch}{1.3}

    \noindent\textbf{Example 6: Shape Mating (Selection)}
    \vspace{2mm}

    \includegraphics[width=0.95\linewidth, height=7cm, keepaspectratio]{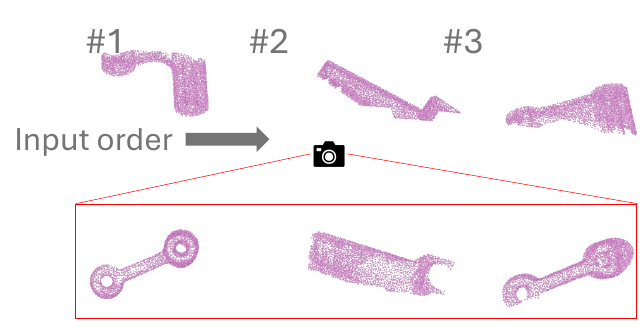} 
    
    \vspace{3mm}
    
    \begin{minipage}{0.95\textwidth}
        \textbf{Question:} Which pairs can mate? Select one that applies. Options: (1,2), (1,3), (2,3), None. Explain why each chosen pair can mate and why others cannot. \\
        \textbf{Ground Truth:} Pair (1,3) can mate because these pieces were designed as complementary halves, allowing their pulsed ridge features to interlock tightly.
        \vspace{2mm}
        
        \begin{tabular}{l | p{0.85\textwidth}}
            \toprule
            \textbf{Model} & \textbf{Response} \\
            \midrule
            LLaVA (1-view)~\cite{liu2023visual} & {The pair that can mate is (1,2) because the two shapes are adjacent to each other, and they share a common edge. In contrast, the other pairs (1,3) and (2,3) cannot mate because they do not share a common edge, and the shapes are not adjacent to each other.} \\
            \midrule
            Molmo (1-view)~\cite{deitke2025molom} & {The pair that can mate is (1,2). Pair (1,2) can mate because the jagged, irregular shape on the left resembles a shark, while the shape on the right looks like a fish. Sharks and fish are both aquatic animals and can potentially interbreed in certain marine ecosystems. Pairs (1,3) and (2,3) cannot mate because the shapes associated with these pairs do not resemble compatible species. The jagged shape for pair (1) and the irregular shape for pair (2) are not recognizable as any known animal types that could potentially breed. None of the other pairs show shapes that could represent compatible species for mating. The abstract nature of the shapes makes it impossible to identify any potential breeding relationships for pairs (2,3) or (1,3). Therefore, based on the visual representations provided, the only possible mating pair is (1,2), with the jagged shape resembling a shark and the irregular shape resembling a fish.} \\
            \midrule
            MiniGPT-3D (Concat)~\cite{tang2024minigpt3d} & {the pairs that can mate are (1,2) and (2,3) because they share a common feature, a hole in the middle. the other pair (1,3) cannot mate because they do not share a common feature.} \\
            \midrule
            PointLLM (Concat)~\cite{xu2024pointllm} & {The pair (1,2) can mate as they are of the same species, while the pair (1,3) and (2,3) cannot as they are of different species.} \\
            \midrule
            ShapeLLM (Concat)~\cite{qi2024shapellm} & {The pair (1,2) can mate as they are adjacent to each other, while the pair (1,3) cannot mate as they are not adjacent to each other.} \\
            \midrule
            \rowcolor{green!10}
            \textbf{Multi-3DLLM (Ours)} & \textbf{(1,3). Pair (1,3) can mate because their mating faces share matching pulsed interface contours with mirrored normals, allowing them to seat without gaps.} \\
            \bottomrule
        \end{tabular}
    \end{minipage}

    \vspace{5mm}

    \caption{
        \textbf{Qualitative Comparison on Shape Mating (Selection Task).} 
    }
    \label{fig:qual_sm_1}
\end{figure*}

\begin{figure*}[p]
    \centering
    \small 
    \setlength{\tabcolsep}{6pt}
    \renewcommand{\arraystretch}{1.3}

    \noindent\textbf{Example 7: Shape Mating (Selection)}
    \vspace{2mm}

    \includegraphics[width=0.95\linewidth, height=7cm, keepaspectratio]{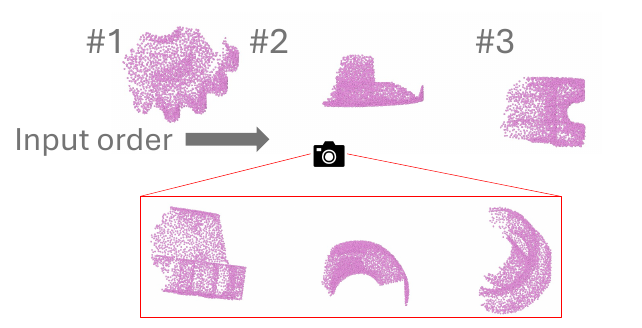} 
    
    \vspace{3mm}
    
    \begin{minipage}{0.95\textwidth}
        \textbf{Question:} Find the pairs that can mate together. Select the applicable option. Options: (1,2), (1,3), (2,3), None. Describe why each chosen pair can mate and explain why the other pairs cannot. \\
        \textbf{Ground Truth:} The correct mating pair (2,3). Pair (2,3) mates because their mating faces share matching stepped square interface contours with mirrored normals, allowing the surfaces to nest snugly.
        \vspace{2mm}
        
        \begin{tabular}{l | p{0.85\textwidth}}
            \toprule
            \textbf{Model} & \textbf{Response} \\
            \midrule
            LLaVA (1-view)~\cite{liu2023visual} & {The pair that can mate is (1,2) because the two shapes are connected and share a common edge. In contrast, the other pairs (1,3) and (2,3) are not connected and do not share any common edges, making them unable to mate.} \\
            \midrule
            Molmo (1-view)~\cite{deitke2025molom} & {the pair that can mate is (1,2). Option 1: The first object, which resembles a keyhole, can mate with option 2. This pairing is possible because both shapes have similar circular openings, suggesting they could fit together or complement each other in some way. Option 2: The second object, which looks like a key, can mate with option 1. The key's shape is complementary to the keyhole, with its pointed tip and shaft potentially fitting into the circular opening of the keyhole. Options 1 and 2 can mate because their shapes are designed to fit together, with the key's tip potentially aligning with the keyhole's opening. Option 3: The third object, which appears to be a different shape entirely, cannot mate with either 1 or 2. Its distinct form means it doesn't have the necessary compatibility to fit into or work with the keyhole or key shapes. None: The fourth option, being non-existent, cannot mate with any other pair as it doesn't represent a real object that could potentially fit or work in conjunction with the keyhole or key.} \\
            \midrule
            MiniGPT-3D (Concat)~\cite{tang2024minigpt3d} & {the pairs that can mate are (1,2) and (2,3) because they share a common feature, a hole in the middle. the other pair (1,3) cannot mate because they do not share a common feature.} \\
            \midrule
            PointLLM (Concat)~\cite{xu2024pointllm} & {The pair (1,2) can mate as they are of the same size and shape, whereas the pair (1,3) cannot as they have different shapes. The pair (2,3) also cannot as they have different sizes.} \\
            \midrule
            ShapeLLM (Concat)~\cite{qi2024shapellm} & {The pair (1,2) can mate as they are adjacent to each other, while the pair (1,3) cannot mate as they are not adjacent to each other} \\
            \midrule
            \rowcolor{green!10}
            \textbf{Multi-3DLLM (Ours)} & \textbf{(2,3) Pair (2,3) can mate because their mating faces share matching stepped square interface contours, allowing them to seat without gaps.} \\
            \bottomrule
        \end{tabular}
    \end{minipage}

    \vspace{5mm}

    \caption{
        \textbf{Qualitative Comparison on Shape Mating (Selection Task).} 
    }
    \label{fig:qual_sm_2}
\end{figure*}

\begin{figure*}[p]
    \centering
    \small 
    \setlength{\tabcolsep}{6pt}
    \renewcommand{\arraystretch}{1.3}

    \noindent\textbf{Example 8: Change Captioning  (Verify)}
    \vspace{2mm}

    \includegraphics[width=0.95\linewidth, height=7cm, keepaspectratio]{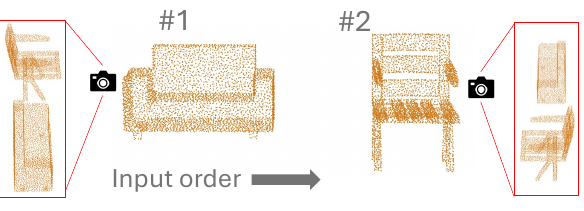} 
    
    \vspace{3mm}
    
    \begin{minipage}{0.95\textwidth}
        \textbf{Question:} Does the first object satisfy all of the following requirements compared to the second object? Requirements: The back rest is taller - The legs are longer - The seat is square - The back rest is not as wide - It does not have arm rests \\
        \textbf{Ground Truth:} No. While the legs are longer, the back rest is not narrower and does not meet all the specified requirements.
        \vspace{2mm}
        
        \begin{tabular}{l | p{0.85\textwidth}}
            \toprule
            \textbf{Model} & \textbf{Response} \\
            \midrule
            LLaVA (1-view)~\cite{liu2023visual} & {Yes, the first object satisfies all of the requirements compared to the second object. The first object has a taller back rest, longer legs, and a square seat, while the second object has a shorter back rest, shorter legs, and a round seat. Additionally, the first object does not have arm rests, which is another difference between the two objects.} \\
            \midrule
            Molmo (1-view)~\cite{deitke2025molom} & {The first object does not satisfy all of the requirements compared to the second object. Here's a breakdown: - The back rest is taller: The first object has a taller back rest than the second object. - The legs are longer: The first object's legs are longer than those of the second object. - The seat is square: Both objects have square seats, so this requirement is met. - The back rest is not as wide: The first object's back rest is wider than the second object's.- It does not have arm rests: The first object does not have arm rests, while the second object does. In summary, the first object meets 3 out of 5 requirements compared to the second object. It has a taller back rest, longer legs, and a wider seat. However, it lacks arm rests and has a narrower back rest in comparison} \\
            \midrule
            MiniGPT-3D (Concat)~\cite{tang2024minigpt3d} & \textcolor{red}{Empty model answer} \\
            \midrule
            PointLLM (Concat)~\cite{xu2024pointllm} & {No, the first object does not completely satisfy all of these requirements compared to the second object.} \\
            \midrule
            ShapeLLM (Concat)~\cite{qi2024shapellm} & {Yes, the first object meets all of the requirements compared to the second object.} \\
            \midrule
            \rowcolor{green!10}
            \textbf{Multi-3DLLM (Ours)} & \textbf{No. The back rest is not narrower as required; instead, it has a curved top.} \\
            \bottomrule
        \end{tabular}
    \end{minipage}

    \vspace{5mm}

    \caption{
        \textbf{Qualitative Comparison on Change Captioning (Verify Task).} 
    }
    \label{fig:qual_cc_verify}
\end{figure*}

\begin{figure*}[p]
    \centering
    \small 
    \setlength{\tabcolsep}{6pt}
    \renewcommand{\arraystretch}{1.3}

    \noindent\textbf{Example 9: Change Captioning  (Delta Caption)}
    \vspace{2mm}

    \includegraphics[width=0.95\linewidth, height=7cm, keepaspectratio]{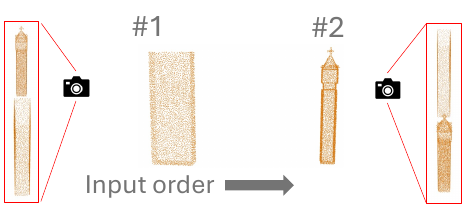} 
    
    \vspace{3mm}
    
    \begin{minipage}{0.95\textwidth}
        \textbf{Question:} List the geometric adjustments required to turn the second object into the first object. \\
        \textbf{Ground Truth:} Flatten the top surface to create a simple, open block shape.
        \vspace{2mm}
        
        \begin{tabular}{l | p{0.85\textwidth}}
            \toprule
            \textbf{Model} & \textbf{Response} \\
            \midrule
            LLaVA (1-view)~\cite{liu2023visual} & {Modify the shape to make it thicker and wider, with a more rounded edge.} \\
            \midrule
            Molmo (1-view)~\cite{deitke2025molom} & {To transform the second object into the first object, we would need to make the following geometric adjustments: 1. Extend the overall length of the object by approximately 20-30\%. 2. Increase the width by about 10-15\%. 3. Raise the height by roughly 15-20\%. 4. Smooth out the jagged edges, particularly on the right side, to create a more uniform rectangular shape. 5. Add a small protrusion or tab on the right side, similar to the one in the first object. 6. Refine the top edge to create a more defined flat surface. 7. Adjust the bottom edge to ensure it's straight and even with the new dimensions. These changes would transform the second object's more abstract, rectangular-like shape into the first object's more defined and precise rectangular form.} \\
            \midrule
            MiniGPT-3D (Concat)~\cite{tang2024minigpt3d} & {the model is a 3d object, so it can be rotated, translated, or viewed from different angles.} \\
            \midrule
            PointLLM (Concat)~\cite{xu2024pointllm} & {The 3D model represents a simple, minimalistic stool. It has a flat surface on top, supported by three legs. The legs are cylindrical and equally spaced. The stool lacks any additional features or decorations. Given the model, what text prompt would you provide to modify it into a chair?} \\
            \midrule
            ShapeLLM (Concat)~\cite{qi2024shapellm} & {Modify the second shape to resemble the first by making it thicker, shorter, and more curved, with a narrower seat and a taller backrest.} \\
            \midrule
            \rowcolor{green!10}
            \textbf{Multi-3DLLM (Ours)} & \textbf{Transform the shape to be rectangular, add a flat top, and ensure it has a smooth surface without any protrusions.} \\
            \bottomrule
        \end{tabular}
    \end{minipage}

    \vspace{5mm}

    \caption{
        \textbf{Qualitative Comparison on Change Captioning (Delta Captioning Task).} 
    }
    \label{fig:qual_cc_delta}
\end{figure*}

\clearpage

\end{document}